\documentclass[lettersize,journal]{IEEEtran}
\usepackage{amsmath,amsfonts}
\usepackage{algorithmic}
\usepackage{algorithm}

\usepackage{array}
\usepackage[caption=false,font=normalsize,labelfont=sf,textfont=sf]{subfig}
\usepackage{textcomp}
\usepackage{stfloats}
\usepackage{url}
\usepackage{verbatim}
\usepackage{graphicx}
\usepackage{cite}
\usepackage{color}
\usepackage[cmyk]{xcolor}
\usepackage{hyperref}
\usepackage{multirow}
\usepackage{threeparttable} 
\makeatletter

\newcommand{\Rmnum}[1]{\expandafter\@slowromancap\romannumeral #1@}
\makeatother

\begin{document}

\title{Multistage Large Segment Imputation Framework Based on Deep Learning and Statistic Metrics}

\author{
	\IEEEauthorblockN{JinSheng Yang\IEEEauthorrefmark{1}, YuanHai Shao\thanks{* YuanHai Shao is the corresponding author.}\IEEEauthorrefmark{1}, ChunNa Li\IEEEauthorrefmark{1}, Wensi Wang\IEEEauthorrefmark{2}}\\
	\IEEEauthorblockA{\IEEEauthorrefmark{1} Management School,, Hainan University, Haikou, 570228, P.R.China}\\
	\IEEEauthorblockA{\IEEEauthorrefmark{2} Faculty of Information Technology, Beijing University of Technology, 12496 Beijing, China}\\
	\thanks{This work is supported by the Natural Science Foundation of Hainan Province (No.120RC449), the National Natural Science Foundation of China (Nos. 61866010, 11871183), the Scientific Research Foundation of Hainan University (No: kyqd(sk)1804), and the Key Laboratory of Engineering Modeling and Statistical Computation of Hainan Province.} 
}


\maketitle

\begin{abstract}
Missing value is a very common and unavoidable problem in sensors, and researchers have made numerous attempts for missing value imputation, particularly in deep learning models. However, for real sensor data, the specific data distribution and data periods are rarely considered, making it difficult to choose the appropriate evaluation indexes and models for different sensors. To address this issue, this study proposes a multistage imputation framework based on deep learning with adaptability for missing value imputation. The model presents a mixture measurement index of low- and higher-order statistics for data distribution and a new perspective on data imputation performance metrics, which is more adaptive and effective than the traditional mean squared error. A multistage imputation strategy and dynamic data length are introduced into the imputation process for data periods. Experimental results on different types of sensor data show that the multistage imputation strategy and the mixture index are superior and that the effect of missing value imputation has been improved to some extent, particularly for the large segment imputation problem. The codes and experimental results have been uploaded to GitHub \footnote{https://github.com/BomBooooo/MLSIF/tree/main}. 
\end{abstract}


\section{Introduction}
\IEEEPARstart{W}{ith} the fast growth of sensor applications, the amount of sensor data has increased tremendously in recent years \cite{chen2014vision}. The missing value has been a major impediment to the development of the sensor data analysis process \cite{stankovic2014research}. Missing data values can be caused by various factors, including sensor failure, data loss, irregular sampling data, manual recording errors, sensor maintenance, and debugging \cite{junninen2004methods}. Missing value is a widespread and difficult problem to avoid; it complicates future data analysis for researchers and engineers. 

Currently, there are primarily two types of methods \cite{lin2020missing} for dealing with missing values, namely, deletion and imputation. Deletion is the act of directly deleting data with missing values; this will not only result in the data loss of specific information but also lead to incomplete time series, thereby affecting the subsequent data analysis work. The imputation methods \cite{fang2020time} are divided into traditional machine learning methods and deep learning methods. Traditional machine learning methods include neighbor-based methods \cite{hudak2008nearest}, constraint-based methods \cite{song2015screen}, regression-based methods \cite{zhang2017time}, statistical-based methods \cite{zhang2016sequential}, matrix factorization-based methods \cite{morup2010scalable}, expectation maximization-based methods \cite{ghahramani1994supervised} and imputation multivariate imputation by chained equation-based methods \cite{van2011mice, azur2011multiple}. These methods are better suited to situations with a small amount of data and a low missing rate. Deep learning methods include fully connected neural network (FCNN)-based methods \cite{yoon2018gain}, convolutional neural network (CNN)-based methods \cite{guo2019data}, recurrent neural network (RNN)-based methods \cite{yoon2017multi, luo2018multivariate, lipton2016modeling, dabrowski2019sequence, cao2018brits, suo2019recurrent, liu2019naomi, mulyadi2021uncertainty, ma2019end}, generative adversarial network (GAN)\cite{luo2019e2gan, gupta2020time, miao2021generative, li2019misgan}, attention \cite{suo2020glima, ma2019cdsa}, and transformer \cite{shan2021nrtsi}. Deep learning methods are better suited with a large amount of data and a high missing rate. 

Although deep learning methods have made significant progress in the problem of missing value imputation, their assumptions and settings are quite different. In practice, the characteristics of the missing values for sensor data are complicated, with factors such as missing rate, missing location, and maximum missing length, all of which will influence the difficulty of the imputation task. Making many assumptions in advance may result in unexpected outcomes that cannot be used in practice. The current deep learning imputation methods have three types of issues.

First, the measurement indexes of missing value imputation warrant further study \cite{junninen2004methods}. To measure the imputation effect of missing values point to point, the current practice is to use the indicators from the mean squared error (MSE) category, particularly, MSE, mean absolute error (MAE), root MAE (RMAE), and root MSE (RMSE). They are supervised indexes, which means that the operation steps are to first remove a subset of the values based on the original data, use the model to impute this subset of the values, and finally use RMSE or MAE to measure the gap between the removed and imputed values. MSE and MAE are mainly used at imputation tasks \cite{junninen2004methods}. But, they overemphasize the distinction between dropped and imputed values while neglecting the difference between the missing and imputed values. Furthermore, when a part of the data is removed in advance, the method cannot be used completely for model training and imputation. This issue is especially prominent when some data has a high missing rate. 

Second, as we all know, when processing a large amount of time-series data in deep learning, the data is divided into small segments, each of which is a subsequence of the original time-series data, also known as a sample in deep learning. As a result of this crucial step, the missing rate can be divided into two categories: local missing and global missing rate, with the global missing rate representing the missing rate of all data and the local missing rate representing the missing rate of each short segment. In essence, the model should account for the local missing rate. The local missing rate for different data segments can range from 0\% to 100\%, depending on the missing rate of the data and the length of the segment. By contrast, the global missing rate is fixed. However, most of the existing deep learning imputation methods \cite{yoon2018gain, guo2019data, yoon2017multi, luo2018multivariate, lipton2016modeling, dabrowski2019sequence, cao2018brits, suo2019recurrent, gupta2020time, miao2021generative, ma2019cdsa, luo2019e2gan, liu2019naomi, shan2021nrtsi, suo2020glima, li2019misgan, mulyadi2021uncertainty, ma2019end} do not distinguish between these two concepts. At this point, two subproblems arise.

\romannumeral1) The imputation task and the model’s target are not aligned. The task that needs to be accomplished is to fill in the missing values in the data. However, it is easy to transform the problem of imputing fixed global missing rates into imputing fixed local missing rates during evaluation. Missing data in actual sensor data easily cluster together; hence, the local missing rate easily becomes zero. 

\romannumeral2) The choice of using disguised missing data is unrealistic. Most of the aforementioned studies only consider nearly or completely complete data on missing value imputation. Removing values with different missing rates makes it easier to train and evaluate the model with more complete data. However, removing and calculating the loss when there are missing values in the sample data is unclear in the training phase; in the testing phase, how to select test data and what indicators to be used to measure the model’s quality are also very sensitive. In practice, the imputed data contains missing values; how to train and evaluate this type of data is a topic worth discussing. 

\begin{figure}[!t]
	\centering
	\includegraphics[width=3.4in]{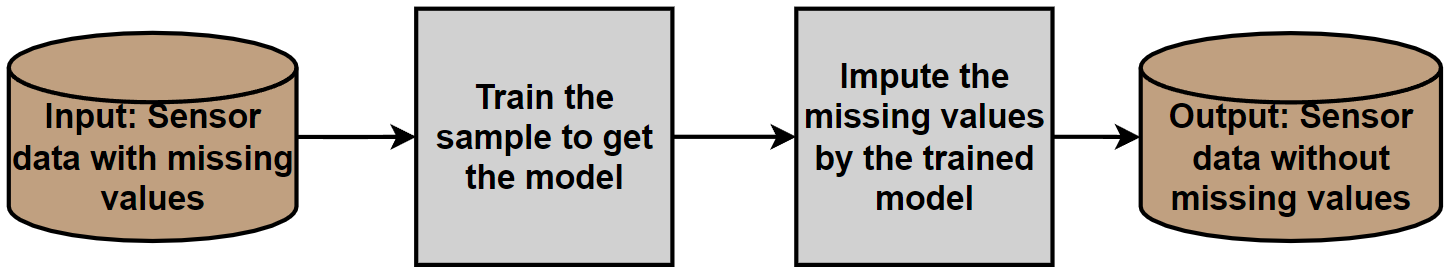}
	\caption{The flowchart of one-stage imputation.}
	\label{one stage}
\end{figure}

Third, more adaptable imputation strategies are required for different missing value problems. However, the conventional approach of treating the missing value imputation as a one-stage problem also overlooks the assumptions of multiple local missing rates. Fig. \ref{one stage} shows a one-stage flowchart. Liu et.al \cite{liu2020missing} proved that the iterative approach improves imputation performance. The missing rate of data should always be kept constant when dealing with data with different missing conditions using the same method. This can be accomplished through an iterative approach. However, most researchers use the strategy of fixed length segmentation processing for long-length sensor data; however, fixed length imputation cannot deal with large missing sections appropriately. \cite{liu2020missing} has proven that RMSE increases concerning the gap under the condition of a fixed sample length, i.e., the larger the missing gap in the data, the more information the model requires. 

To address the aforementioned issues, we propose a Multistage Large Segment Imputation Framework (MLSIF) in this study. The MLSIF introduces a new statistical indicator that allows missing data values to participate in the entire stage of the training and evaluation process, allowing the MLSIF to impute sensor data with missing data. To improve the imputation effect, the framework design uses an iterative multistage imputation method and uses different data segment lengths in each imputation stage to better process different missing conditions. The real sensor data is used in the experimental part to verify the model's quality, and by simulating the missing situation of real data, the problem of separation from the real situation caused by artificial regulations is avoided. The deep learning model, NRTSI \cite{shan2021nrtsi}, is used as the basic imputation model structure of the framework. The main contributions of this study are summarized as follows. 

\begin{itemize}
	\item{A new statistical indicator is presented. The newly proposed indicator can include the missing values from the data itself in the entire training and testing process and can be generated based on the distribution of the data without removing the original data. Therefore, it is more consistent with real-world situations and can better measure the effect of imputation to a certain extent.}
	
	\item{A new method of constructing missing values is adopted, i.e., using complete or nearly complete data to simulate the missing situation of real data to avoid the deviation from the actual situation caused by an artificial setting.}
	
	\item{A new MLSIF is proposed that can handle the missing value imputation task for actual missing data. The large-missing gaps are divided into different stages and imputed among the multistage. MLSIF dynamically changes the length of training and imputing data based on the missing situation of the current data, and longer missing gaps are provided with longer observation data, which means more information. Hence, the real-life sensor data can be handled.}
	
	\item{The experimental results on both the benchmark and real sensor data show the effectiveness of the proposed MLSIF. The superiority of the multistage imputation strategy and the mixture loss have been highlighted, and the effect of missing value imputation has been improved to some extent, especially for the large missing gap imputation problem.}
\end{itemize}

\section{Problem Formulation}

This study considers the problem of missing value imputation in univariate sensor data, a class of time series data. Let $\bold X=\{x_1,x_2,x_3,…,x_t\}$ be a sequence of sensor data with length $t$, where $x_t \in \mathbb{R}$, and some $x_i$ may be missing. To identify the missing values, define a mask sequence $\bold M=\{m_1,m_2,m_3,…,m_t\}$ that corresponds to $\bold X$, where 

\begin{equation}
	{m_j} = \begin{cases}
		0,&{\text{if}}\ x_j {\text{ is missing}}\\ 
		{1,}&{\text{otherwise,}} 
	\end{cases}
	\label{mask}
\end{equation}
$j=1,2,\ldots,t$. Denote $\bold {x_i^L}=\{x_i,x_{i+1},x_{i+2}, ...,x_{i+L-1}\}$ as a sample, which is a subsequence of $\bold X$, where $L \in \mathbb{N}^* $ and $ \mathbb{N}^* $ are the set of positive integers. The problem we need to solve is how to impute the missing part.

Consequently, $\bold X$ can be decomposed into multiple subsequences with length $L$ without crossover. Define a set $\bold {X^{L}_{split}}=\{\bold {x_i^L}\}_{i=1}^N$ that contains all the subsequences of $\bold X$, where $N$ represents the number of samples. When the data cannot be completely segmented, the last few data will form an $ \bold {x_N^L} $ with the previous consecutive data. Thus $ N $ can be calculated using Equation (\ref{N}).

\begin{equation}
		{N} = \begin{cases}
		{\frac{t}{L}},&{\text{if}}\ \frac{t}{L} \in \mathbb{N}^*\\ 
		{[\frac{t}{L}] + 1,}&{\text{otherwise,}} 
	\end{cases}
	\label{N}
\end{equation}
where $[\cdot]$ is least integer function. 

\section{Related Works}

\subsection{Deep Learning Models}
Recently, there have been several studies on imputation deep learning models in the field of missing value imputation. It should be noted that sensor data is a type of time-series data. Therefore, time-series models can be applied to sensor data. 

FCNN, RNN, and CNN were applied as the three basic structures in the early days of neural network development. As the first proposed network structure, FCNN can achieve good results while performing several tasks, including missing value imputation \cite{yoon2018gain}. However, due to its insensitivity to time and a large number of parameters, it is gradually being replaced in time-series data by RNN and in image data by CNN. RNN is frequently mentioned by researchers in missing value imputation tasks as a network structure designed for time-series data, such as unidirectional RNN \cite{suo2019recurrent, lipton2016modeling, shen2018end}, bidirectional RNN \cite{cao2018brits, dabrowski2019sequence, zhang2019ssim}, and variant RNN \cite{yoon2017multi, liu2019naomi}. CNN has achieved unparalleled results in image processing; however, it has not shown unique advantages in time series-data. Guo \cite{guo2019data} attempted to use CNN for missing value imputation and obtained acceptable results.

The breadth and depth of neural networks have steadily increased with the exponential growth of data and processing throughout this century. Some new network, such as GAN and attention, have gradually emerged. GAN has shown promising results in image generation; thus, some researchers have attempted to apply it for missing value generation \cite{luo2018multivariate, luo2019e2gan, gupta2020time, miao2021generative, li2019misgan}. Attention is a model designed for language sequence data that can also be used for missing value imputation in time-series data. GLIMA \cite{suo2020glima} believes that the information between the part and the whole data should be fully considered; therefore, it constructs a structure that can not only extract local information but also consider the whole information. Ma \cite{ma2019cdsa} used attention to extract information from data to realize missing value imputation. 

Based on the attention structure, the transformer structure begins to show better results on a wide range of tasks. It is possible to think of it as a compound neural network structure based on attention. NRTSI \cite{shan2021nrtsi} uses a nonregression method to impute the missing values in the field of missing value imputation. It reinterprets time series as a set of (time, data) tuples and proposes a time-series imputation method based on a permutation equivariance model, achieving excellent results so far in time-series imputation experimental results. 

\subsection{Imputation Frameworks}
Compared with imputation models that use processed sequences and samples, a more integral and comprehensive approach is to use a strategic framework to complete the imputation, which refers to the entire imputation process from missing data to complete data. 

Some researchers have attempted to solve the problem of missing value imputation using a framework. Farhangfar \cite{farhangfar2007novel} provided a framework that can be used in almost any method to generate weights representing the quality of each estimate to perform boosting. Applying it to an imputation method can, on average, significantly improve the imputation accuracy while maintaining the same asymptotic computational complexity. Rahman \cite{rahman2014fimus} subsequently proposed a framework for imputing missing values based on coappearance, correlation and similarity analysis. It proposes a novel missing value imputation technique that uses existing dataset patterns such as co-occurrence of attribute values, correlations between attributes, and similarity of attribute values.

Although some general frameworks exist, there are some flaws for specific problems, such as their inability to distinguish between local and global missing rates and their poor performance on the large gap missing problem. 

\subsection{Performance Metrics}
Metrics are the criteria by which a model is measured, evaluated and selected in the task of imputing missing values. There are three commonly used metrics \cite{junninen2004methods}: MSE-like evaluation (MSE, MAE, RMSE, and RMAE), $d_2$ and $R^2$. Assume two sequences of time series data $\bold P = \{p_1, p_2, ..., p_n\}$ and $\bold O = \{o_1, o_2, ..., o_n\}$ of equal length, where $n \in \mathbb{N}^*$. Let $\bold O$ be the obvious values removed artificially and $\bold P$ be the imputed values of the corresponding position. MSE and MAE can be defined using Equations (\ref{MSE}) and (\ref{MAE}). 

\begin{equation}
	MSE(\bold P, \bold O) = \frac{1}{n}\sum_{i=1}^n|p_i - o_i|^2
	\label{MSE}
\end{equation}
\begin{equation}
	MAE(\bold P, \bold O) = \frac{1}{n}\sum_{i=1}^n|p_i - o_i|
	\label{MAE}
\end{equation}
 RMSE and RMAE are the root of MSE and MAE, respectively. Moreover, $d_2$ and $R^2$ are defined as follows:

\begin{equation}
	d_2(\bold P, \bold O) = 1 - (\frac{\sum_{i=1}^n(p_i - o_i)^2}{\sum_{i=1}^n(|p_i - \bar{o}| + |o_i - \bar{o}|)^2})
	\label{d_2}
\end{equation}

\begin{equation}
	R^2(\bold P, \bold O) = (\frac{1}{n}\frac{\sum_{i=1}^n[(p_i - \bar{p}) * (o_i - \bar{o})]}{\sigma_p \sigma_o})^2
	\label{R^2}
\end{equation}
where $\bar{o}, \bar{p}, \sigma_o,$ and $\sigma_p$ are the mean and standard deviation of $\bold O$ and $\bold P$.

To use the above indicators, missing values must be artificially created in nonmissing value data. Specifically, first, remove a portion of the values from the data and subsequently impute the removed values. Furthermore, the imputation effect is measured by comparing the difference between the removed values and the imputed values. 

MSE-type metrics have a higher tolerance for results that are close to the mean; however, they are sensitive to extreme values, resulting in less attention being given to the overall data distribution. $d_2$ and $R^2$ are commonly used in statistical analysis and focus on the difference between the total data and the mean, which is the measure of dispersion. More importantly, the above metrics must have corresponding true values before they can be calculated, i.e., one must first know the corresponding original value before measuring it. This requirement is unrealistic in practice. When the rate of missing data is low, it can be used to assess how effectively the imputed values are performing. However, when the missing rate of the data is high, the artificially constructed missing values will jeopardize the data's integrity and reduce the amount of valid information in the data. 

\subsection{Imputation Losses}
In practice, the loss directs the model training. However, it is difficult to directly find computable targets for most tasks in practice; hence approximating methods are used to achieve the goal. The goal of the missing value imputation task is to minimize the expectation of loss between missing and imputed values in Equation (\ref{impute objective}).
\begin{equation}
	\underset {\theta} {\text{min}}\ E\ (\mathcal{L} (X_{real} \odot (1-M),\ F(X,\theta) \odot (1-M)))
	\label{impute objective}
\end{equation}
where $\mathcal{L} (\cdot)$ represents a loss function, $X_{real} \in \mathbb{R}^t$ is the complete time-series data, $\odot$ represents element-wise multiplication and $F$ is an imputation model with parameter $\theta$. $X_{real} \odot M$ denotes the observed portion of $X_{real}$ and $X_{real} \odot (1-M)$ denotes the missing portion of $X_{real}$.

This task is difficult to achieve, because $X_{real}*(1-M)$ is not known. Therefore, the traditional regression method sets an approximate loss construction method shown in Equation (\ref{regression loss}). 
\begin{equation}
	L_{regression}=\underset {\theta} {\text{min}} \parallel X_{real} \odot M - F(X,\theta) \odot M \parallel ^2
	\label{regression loss}
\end{equation}
where $\parallel \cdot \parallel$ represents the two-norm.

However, T.M. Choi \cite{choi2020rdis} emphasized that the loss determined by the traditional regression method does not correspond to the task to be completed. The difference between the regressed and observed values is calculated using Equation (\ref{regression loss}). There is an implicit assumption in this loss that when the imputed values are close enough to the real values, the imputation result is satisfactory. Although this loss is simple to calculate, it differs significantly from Equation (\ref{impute objective}). It calculates the regression loss rather than the imputation loss, which does not match the target well. Therefore, training imputation networks with Equation (\ref{regression loss}) can be called implicit training. For these reasons, T.M. Choi \cite{choi2020rdis} proposed a new training method for explicit training based on random drop imputation with self-training (RDIS). Random drop data are generated by randomly removing existing values in the time-series data $\bold X$. The random drop data is denoted as $\bold {\check{X}}=(\check{x}_1,\check{x}_2,\check{x}_3,...,\check{x}_t)$ and $\bold {\check{M}}=(\check{m}_1,\check{m}_2,\check{m}_3,...,\check{m}_t)$, where
\begin{equation}
	{\check{m}_t} = \begin{cases}
		1,&{\text{if}}\ \check{x}_t {\text{ is droped}}\\ 
		{0,}&{\text{otherwise.}} 
	\end{cases}
\end{equation}

The loss function of RDIS can be expressed as follows:
\begin{equation}
	\begin{aligned}
		L_{RDIS} = \underset {\theta} {\text{min}}\ (&\parallel X \odot (M-\check{M})-F(\check{X},\theta) \odot (M-\check{M}) \parallel ^2 \\
		&+ \parallel X \odot \check{M}-F(\check{X},\theta) \odot \check{M} \parallel ^2)
	\end{aligned}
	\label{rdi loss}
\end{equation}
where $X \odot \check{M}$ represents the dropped part of the observations, and $X \odot (M-\check{M})$ represents the remaining part of the observations after dropping.

Compared with Equation (\ref{regression loss}), Equation (\ref{rdi loss}) is more advanced, where it adds a second section to the loss function, making it more similar to Equation (\ref{impute objective}). It removes the data and converts the information that will be used as input to the model. This approach is useful for model training when the sample missing rate is low. When the sample missing rate is high, the information of the input data is further destroyed. 

Compared to the regression loss, the nonregression method \cite{liu2019naomi, shan2021nrtsi} only outputs the missing position values. Therefore, the loss can only be calculated by artificially removing certain observed data. The loss function is shown in Equation (\ref{non-regression loss}).
\begin{equation}
	L_{non-reg} = \underset {\theta} {\text{min}}\ (\parallel X \odot \check{M}-F(\check{X},\theta) \odot \check{M} \parallel ^2)
	\label{non-regression loss}
\end{equation}

The above loss (\ref{non-regression loss}) is a step closer to the objective Equation (\ref{impute objective}) than the Equation (\ref{rdi loss}), which eliminates the influence of the observation data on the loss, and guides model training by directly calculating the loss of the missing data. However, all of these losses are faced with two problems. First, when the sample missing rate is high, adding missing values degrades the original data. Second, when MSE-like loss is used as a loss function to guide model training, the imputation result will be very close to the mean because of its sensitivity to extreme values \cite{junninen2004methods}.

Overall, the four aspects of work mentioned above are crucial for the missing value imputation task. The effect of imputation is determined by model and loss, the framework by how imputation is performed, and metrics by how good the imputation is. The framework, in particular, controls the input and output, the model imputes the input data and outputs the result, the loss guides the model’s training direction, and the metric is the standard for measuring the imputation quality. The four aspects are interconnected and independent of each other. Improvements in any of these four areas may benefit the missing value imputation task. This study improves and enhances the three problems in the section \uppercase\expandafter{\romannumeral1} from the perspectives of loss function and metrics, experimental design, and framework.

\section{Statistical Indexes Variation Loss and Evaluation Indicator}
This section proposes a statistical indicator that can be calculated directly on the original data with missing values. This is referred to as statistical indexes variation (SIV). SIV, like the MSE-type indicator, could be used both as a loss to guide the model training as well as an evaluation index to assess the quality of imputation results. 

There are three types of assumptions for missing values, \cite{rubin1976inference, hallaji2021dlin}: MCAR, missing at random, and missing not at random. It is difficult to say which type of assumptions are appropriate for real sensor data with missing values. However, the statistical characteristics of the time-series data with and without missing values can be calculated and compared, and these characteristics should not differ significantly when a small amount of data is missing. Consequently, we begin by presenting four statistical indexes: mean, standard deviation, skewness, and kurtosis, each of which indicates distinct data distribution characteristics. Then, SIV is used to calculate the change in statistical indexes before and after data imputation. 

\subsection{Statistical Indexes}
For any given sequence $\bold X$, its statistical characteristics, such as mean ($\mu$) and standard deviation ($\sigma$), the raised power of the corresponding order skewness (S), and kurtosis (K), are calculated using Equations (\ref{mu} - \ref{K}). 
\begin{equation}
	\mu(\bold X) = \frac{\sum_{i=1}^n x_i}{n}
	\label{mu}
\end{equation}
\begin{equation}
	\sigma(\bold X) = \sqrt[2]{\frac{\sum_{i=1}^n (x_i - \mu)^2}{n}}
\end{equation}
\begin{equation}
	S(\bold X) = \sqrt[3]{\frac{1}{n} \sum_{i=1}^n(\frac{x_i - \mu}{\sigma})^3}
\end{equation}
\begin{equation}
	K(\bold X) = \sqrt[4]{\frac{1}{n} \sum_{i=1}^n(\frac{x_i - \mu}{\sigma})^4}
	\label{K}
\end{equation}
where $x_i$ is the element in $\bold X$, $i=1,...,n$.

The mean ($\mu$) describes the middle point of the sample set, and the standard deviation ($\sigma$) describes the average of the Euclidean distances between each sample point in the sample set and the mean. The skewness (S) indicates that a distribution “leans” one way or the other and has an asymmetric tail \cite{cain2017univariate}. This is the amount of data distributed on both sides of the distribution center. The sample data becomes more biased to the right when the skewness is positive. Nevertheless, when the skewness is negative, the sample data becomes more biased to the left. Kurtosis (K) is associated with the distribution's tail, shoulder, and peak \cite{cain2017univariate}. Generally, the smaller the kurtosis, the flatter the data distribution, and the greater the kurtosis, and the more concentrated the data distribution. Skewness and kurtosis, however, can be thought of as the second- and third-order distances from each sample point in the sample set to the mean. We increase S and K to the power of the corresponding order to unify the dimensions. 

\subsection{SIV}
Assuming there are two sequences $\bold X_1$ and $\bold X_2$, the SIV is calculated using Equation (\ref{SIV}). 
\begin{equation}
	SIV(\bold {X_1}, \bold {X_2}) = \mu^{V}_{(\bold {X_1}, \bold {X_2})} + \sigma^{V}_{(\bold {X_1}, \bold {X_2})} + S^{V}_{(\bold {X_1}, \bold {X_2})} + K^{V}_{(\bold {X_1}, \bold {X_2})}
	\label{SIV}
\end{equation}
where $\mu^{V}_{(\bold {X_1}, \bold {X_2})}, \sigma^{V}_{(\bold {X_1}, \bold {X_2})}, S^{V}_{(\bold {X_1}, \bold {X_2})}, K^{V}_{(\bold {X_1}, \bold {X_2})}$ can be obtained by Equation (\ref{change}).

\begin{equation}
	\begin{aligned}
		\mu^{V}_{(\bold {X_1}, \bold {X_2})}=& (\mu(\bold X_1)-\mu(\bold X_2))^2 \\
		\sigma^{V}_{(\bold {X_1}, \bold {X_2})}=& (\sigma(\bold X_1)-\sigma(\bold X_2))^2 \\
		S^{V}_{(\bold {X_1}, \bold {X_2})}=& (S(\bold X_1)-S(\bold X_2))^2 \\
		K^{V}_{(\bold {X_1}, \bold {X_2})}=& (K(\bold X_1)-K(\bold X_2))^2 \\
	\end{aligned}
	\label{change}
\end{equation}

Notably, SIV ssesses the differences in statistical features between the two sequences. SIV has no requirement for the sequence length; it can measure two sequences of different lengths. Thus, it can be used to directly measure the difference between missing sequences before and after imputation. SIV can be used to the missing value imputation problem denoted by Equation (\ref{SIV loss}).
\begin{equation}
	\begin{aligned}
		SIV(\bold {X_{obs}}, \bold {X_{imp}}) = &\mu^{V}_{(\bold {X_{obs}}, \bold {X_{imp}})} + \sigma^{V}_{(\bold {X_{obs}}, \bold {X_{imp}})} \\
		&+ S^{V}_{(\bold {X_{obs}}, \bold {X_{imp}})} + K^{V}_{(\bold {X_{obs}}, \bold {X_{imp}})}
	\end{aligned}
	\label{SIV loss}
\end{equation}
where $\bold {X_{obs}}$ and $\bold {X_{imp}}$ are the observed and completed sequences following imputation, respectively. This equation can be applied as a loss in training objective functions as well as an evaluation index in model selection. 

The SIV loss function is notable for its ability to be calculated directly on original and imputed data. Furthermore, rather than calculating missing values point to point, SIV considers the data distribution characteristics segment by segment. However, it should be noted that these statistical indicators neglect the time information in time-series data. Simultaneously, when the missing data rate is low and the number of imputed values is relatively low, it is simple to validate the effectiveness of SIV. When the missing rate of data is high, it is unclear if the statistical indicators before and after imputation will differ significantly. 

SIV, as an evaluation indicator, can reflect the quality of the imputation effect to some extent. In particular, we present two SIV indexes in experiments. The first calculates the SIV of the overall data before and after imputation using all data as an object; the second calculates and sums the SIV of each piece of data before and after imputation using each piece of data as an object. The first result is referred to as ” Global SIV”, and the value obtained by the second is referred to as ” Local SIV.” 

The SIV proposal addresses the issue of MSE being overused in the task of missing value imputation. On the one hand, SIV can be used as a loss to participate in the model's training. SIV, on the other hand, can assess the quality of missing value imputation from a certain perspective. The most important aspect is that SIV can make the missing values in the data participate in the model's training and evaluation. 

\section{Multistage Imputation Framework}

\begin{figure*}[!t]
	\centering
	\includegraphics[width=7in]{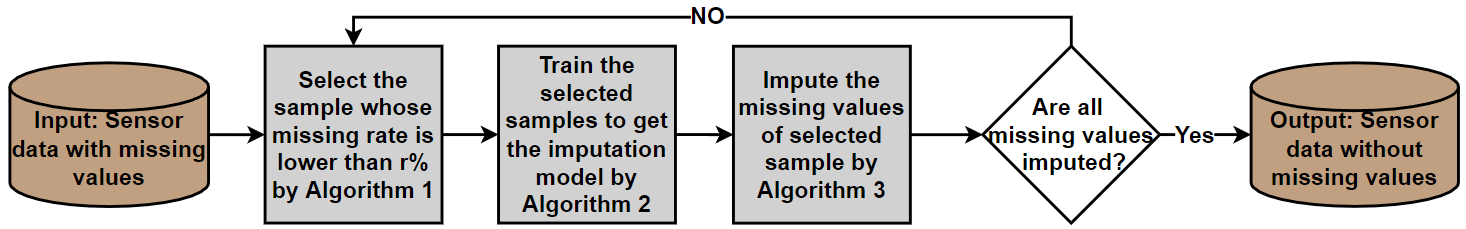}
	\caption{The flow chart of MLSIF.}
	\label{process}
\end{figure*}

In this section, we present MLSIF for the missing value imputation problem. MLSIF adds a cyclic process of selecting and imputing the data to the single-stage imputation method. MLSIF, in particular, employs multistage and dynamic data length tricks. Multistage is reflected in the cyclic structure of the framework, and each cycle is a stage. Each stage imputes the data while keeping the missing rate lower than $r\%$. The term dynamic data length means that the data length will change at each stage. The MLSIF flowchart is shown in Fig. \ref{process}. 

Previous experimental results in the literature \cite{farhangfar2007novel} show that as the missing rate of data increases, the imputation accuracy decreases. Furthermore, Liu \cite{liu2020missing} demonstrated that the iterative approach improves imputation performance. Therefore, iterative multistage is used to echo the dynamic data length. It can alleviate the problem of insufficient effective information caused by large-missing gaps, which makes imputation difficult, if not impossible. Additionally, the data imputed at each stage is the simplest data to impute. Therefore, the goal of MLSIF using dynamic length is to keep the missing rate at a low level. 

In MLSIF, a mixture loss, combined with MSE and SIV, is used to guide the model training. This is because using only the MSE loss causes the model to easily cluster imputed values around the mean, whereas using only SIV causes the loss to be unable to capture temporally characterized values. Consequently, the final result is randomly distributed within the data distribution range with no regularity. 

\subsection{Framework Process}

In each stage in MLSIF, the missing data are imputed once, and each stage contains the four steps described
below. 

\subsubsection*{\bf Step 1: Select the samples whose missing rate is lower than $r\%$ by Algorithm  \ref{alg1}}

The goal of this step is to select high-quality samples for subsequent training. The missing rate $r\%$ is introduced to make it easier to impute the selected samples. The samples with missing rates less than $r\%$ are selected.

\begin{algorithm}[H]
	\caption{Select the samples whose missing rate is lower than $r\%$.}
	\begin{algorithmic}[1]
		\REQUIRE The data: $\bold {X}$; Splitting lengths parameter: $l$; \\
		Missing rate threshold: $r\%$.
		\ENSURE Selected samples set $\bold {X^{L}_{train}}$
		\STATE $L = 0$.
		\STATE Initialize an empty set $\bold  {X^{L}_{train}}$. 
		\STATE \textbf{While} the samples in $\bold {X^{L}_{train}}$ without missing values:
		\STATE \hspace{0.5cm} $L = L+l$.
		\STATE \hspace{0.5cm} Split data $\bold X$ as $\bold {X^{L}_{split}}$. 
		\STATE \hspace{0.5cm} $\bold  {X^{L}_{train}} = \phi$.
		\STATE \hspace{0.5cm} \textbf{for} $\bold {x^{L}}$ in $\bold {X^{L}_{split}}$:
		\STATE \hspace{1.0cm} \textbf{if} the missing rate of $\bold {x^{L}}$ is lower than $r\%$:
		\STATE \hspace{1.5cm} Add $\bold {x^{L}}$ to set $\bold {X^{L}_{train}}$.
	\end{algorithmic}
	\label{alg1}
\end{algorithm}

In Algorithm \ref{alg1}, the input is data $\bold {X} \in \mathbb{R}^t$ with missing values, and the output is a set of samples (segment of X) with a missing rate less than $r\%$, called $\bold {X^{L}_{train}}$. Additionally, there are also two hyperparameters that must be predetermined: $l \in \mathbb{N}^*$ and $r \in [0,100]$. First, initialize the splitting length $L$ to 0 and create an empty set $\bold {X^{L}_{train}}$ as a container for samples with a missing rate less than $r\%$. By increasing the splitting length $L=L+l$, the data $\bold {X}$ is divided into small segments according to the length $L$, called samples. $\bold {X^{L}_{split}}$ is the set of all samples. Set the variable $\bold {X^{L}_{train}} = \phi$. As the set $\bold {X^{L}_{train}}$ is initially empty, this step is only relevant after the second iteration of the loop. Subsequently, iterate over all samples and add those with a missing rate less than $r\%$ to the $\bold {X^{L}_{train}}$. The loop ends only when there are missing values in set $\bold {X^{L}_{train}}$. 

Step 1 involves selecting samples with no missing values and samples with a missing rate less than $r\%$. All of these samples are safer to train and easier to impute than using all samples or the samples with a missing rate greater than $r\%$. 

\subsubsection*{\bf Step 2: Train the selected samples to get the imputation model by Algorithm \ref{alg2}}

The goal of this step is to train the imputation model using the samples selected in step 1 and the MSE + SIV loss.

In Algorithm \ref{alg2}, input the training data set $\bold {X^{L}_{train}}$ selected in Step 1, the deep learning model for imputation $F(\cdot,\theta)$, where NRTSI\cite{shan2021nrtsi} is selected as the imputation model. Additionally, the model training epoch $E$ and the parameters $\gamma \in [0,1], \lambda \in [0,1], \alpha \in [0,1]$ are required. In this equation, $\gamma$ represents the proportion of values dropped during training, $\lambda$ represents the proportional relationship between the imputed and observed values in the $MSE_{loss}$, and $\alpha$ represents the proportional relationship between $MSE_{loss}$ and the $SIV_{loss}$ in the model $L_{MSDIF}$. 

In this step, we use the model to learn about the data’s characteristics. The SIV and MSE coexist. Therefore, we combine SIV and MSE to form the model's loss to guide model training. The model's loss is defined in Equation (\ref{model loss}).
\begin{equation}
	L_{MLSIF}=\underset {\theta} {\text{min}} (1-\alpha)*MSE_{loss}+\alpha*SIV_{loss}
	\label{model loss}
\end{equation}
where hyperparameter $\alpha$ represents the weight between the MSE and SIV. 

This algorithm contains two nested loops. The first layer of loops represents the number of the model training iterations, and the second layer represents the traversal of the training set. The first step of each sample's operation drops a portion of the ample's values based on the proportion of $\gamma$. The dropped data is denoted by $\bold {x_{drop}}$, whereas the remaining data is denoted by $\bold {x_{obs}}$. As the results of previous stages are retained in subsequent stages, $\bold {x_{drop}}$ can be further subdivided into $\bold {x_{drop_{imp}}}$ and $\bold {x_{drop_{obs}}}$, depending on whether the dropped data is imputed or observed data. The model can then be used to impute the missing values, which are denoted as $\bold {x_{predict}}$. Corresponding to the positions of $\bold {x_{drop_{imp}}}$ and $\bold {x_{drop_{obs}}}$, $\bold {x_{predict}}$ is divided into $\bold {x_{predict_{imp}}}$ and $\bold {x_{predict_{obs}}}$. The purpose of splitting the output is to minimize error propagation. There is an error with each imputation iterations, so for multistage imputation, the weights of the original and imputed values should not be the same. $\lambda$ is to control the weight between the two parts. Finally, update model the parameters $F(\cdot,\theta)$ by minimizing $L_{MLSIF}$, where the loss is calculated using the formula given in Equation (\ref{model loss}). 

The innovation of this step is primarily the selection of the loss function. We combine MSE and SIV as the model loss, allowing them to all adhere to their respective strengths. 

\begin{algorithm}[H]
	\caption{Train the imputation model.}
	\begin{algorithmic}[1]
		\REQUIRE The selected samples set: $\bold {X^{L}_{train}}$; \\
		Initialized deep learning imputation model: $F(\cdot,\theta)$; \\
		The epoch of model training: $E$; \\
		The non-negative parameters: $\gamma, \lambda, \alpha$.
		\ENSURE Trained imputation model $F(\cdot,\theta^*)$
		\STATE \textbf{for} $e$ in $range(E)$:
		\STATE \hspace{0.5cm} \textbf{for} $\bold {x^{L}_{train}}$ in $\bold {X^{L}_{train}}$:
		\STATE \hspace{1cm} Random drop $num_{obs}*\gamma$ values in $\bold {x^{L}_{train}}$, where\\
		~~~~~~~~~$num_{obs}$ is the number of the known values in \\
		~~~~~~~~~$\bold {x^{L}_{train}}$.
		\STATE \hspace{1cm} Denote drop values as $\bold {x_{drop}}$ and data remained \\
		~~~~~~~~~as $\bold {x_{obs}}$. In particular, denote the drop data \\
		~~~~~~~~~imputed at previous stages as $\bold {x_{drop_{imp}}}$ and the \\
		~~~~~~~~~drop data from obvious as $\bold {x_{drop_{obs}}}$.
		\STATE \hspace{1cm} $\bold {x_{predict}} = F(\bold {x_{obs}}, \theta)$
		\STATE \hspace{1cm} Corresponding to the positions of $\bold {x_{drop_{imp}}}$ \\
		~~~~~~~~~and $\bold {x_{drop_{obs}}}$, $\bold {x_{predict}}$ is divided into \\
		~~~~~~~~~$\bold {x_{predict_{imp}}}$ and $\bold {x_{predict_{obs}}}$. 
		\STATE \hspace{1cm} Update the parameters of model $M_\theta$ by\\
		~~~~~~~~~minimizing $L_{MLSIF}$ calculated by Equation \\
		~~~~~~~~~(\ref{model loss}), where $MSE_{loss} = $\\
		~~~~~~~~~~~~~~~~~~~~~~$\lambda*MSE(\bold {x_{drop_{obs}}},\bold {x_{predict_{obs}}})+$\\
		~~~~~~~~~~~~~~~~$(1-\lambda)*MSE(\bold {x_{drop_{imp}}},\bold {x_{predict_{imp}}})$,\\
		~~~~~~~~~$SIV_{loss} = SIV(\bold {x^{L}_{train}}, \bold {x_{obs}} \cup \bold {x_{predict}})$.
	\end{algorithmic}
	\label{alg2}
\end{algorithm}

\subsubsection*{\bf Step 3: Impute the missing values of the selected sample by Algorithm \ref{alg3}}

The goal of this step is to use the trained model to impute the missing values in the training data. 

\begin{algorithm}[H]
	\caption{Impute missing values in training data.}
	\begin{algorithmic}[1]
		\REQUIRE Trained imputation model: $F(\cdot,\theta^*)$; \\
		The selected samples set: $\bold {X^{L}_{train}}$.
		\ENSURE $\bold {X^{L}_{train}}$ without missing values
		\STATE \textbf{for} $\bold {x^{L}_{train}}$ in $\bold {X^{L}_{train}}$:
		\STATE \hspace{0.5cm} \textbf{if} $\bold {x^{L}_{train}}$ has missing values:
		\STATE \hspace{1cm} $\bold {x_{predict}} = F(\bold {x^{L}_{train}},\theta^*)$
		\STATE \hspace{1cm} Replace missing values in $\bold {x^{L}_{train}}$ with $\bold {x_{predict}}$. 
	\end{algorithmic}
	\label{alg3}
\end{algorithm}

In Algorithm \ref{alg3}, the model $F(\cdot,\theta^*)$ and the selected samples set $\bold {X^{L}_{train}}$ are input. Use the model to impute the missing values for the samples in $\bold {X^{L}_{train}}$. Finally, the training data is obtained with no missing values. However, outside of the algorithm, the imputed data will replace the missing values in the corresponding missing positions of the original data for later imputation stages. 

\subsubsection*{\bf Step 4: Are all missing values imputed?}

In this step, determine whether all missing values have been imputed and subsequently decide whether to continue the loop. If yes, output the result and end. If no, proceed to Step 1. 

\subsection{Algorithm Summary and Example}

MLSIF investigates flexible imputation strategies. Its entire process constitutes the four steps listed above, which are all interconnected. Step 1 selects data for Steps 2 and 3, and Step 2 can only train the model on that data. The corresponding model in Step 3 is trained using the training data, and only the missing values in the training data can be imputed, i.e., the first step serves as the foundation for all subsequent steps, and both the training and imputing steps are required. 

Overall, there is a strong link between the various stages. Algorithm \ref{alg2} states that the values imputed at previous stages will be provided as information for subsequent stages. This is because imputation becomes more difficult as the number of missing values increases. When the number of missing values decreases, the difficulty of imputation decreases; hence, we begin with low-difficulty tasks first while also providing more information for later high-difficulty imputation tasks; these are the advantages of the multistage and dynamic lengths in MLSIF. 

\begin{figure}[!t]
	\centering
	\subfloat[Step 1 of Stage 1]{
		\begin{minipage}[!t]{0.5\textwidth}
			\centering
			\includegraphics[width=3.4in]{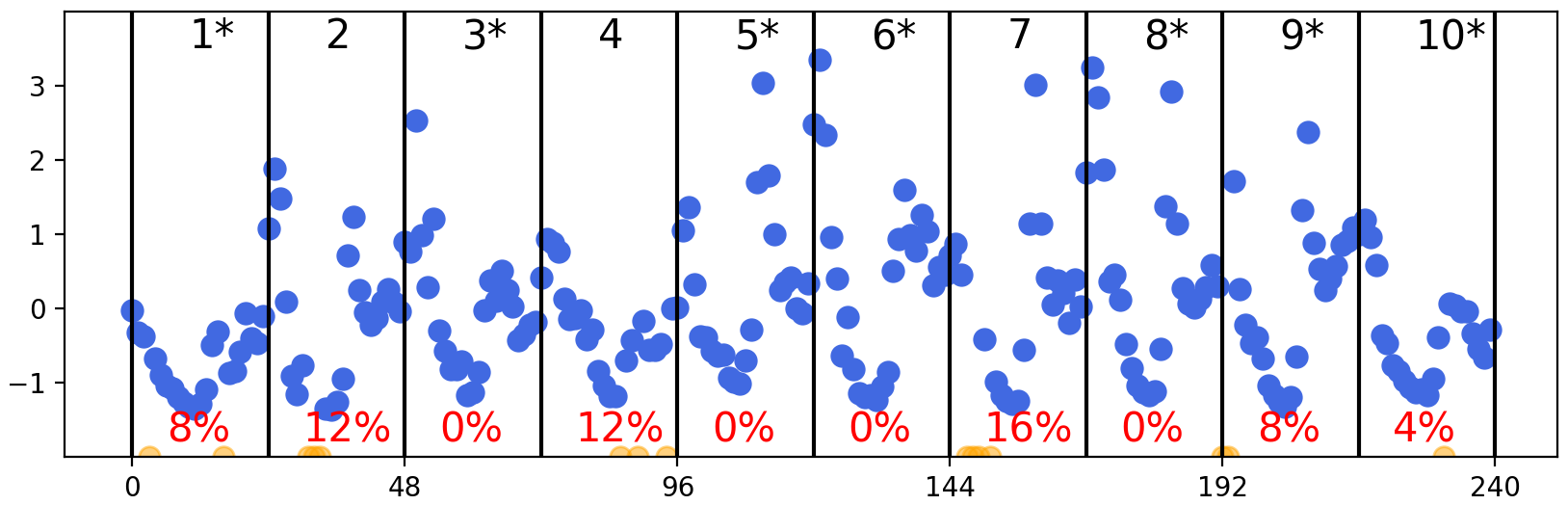}
			\label{split1}
		\end{minipage}
	}
	\\
	\subfloat[Result of Stage 1]{
		\begin{minipage}[!t]{0.5\textwidth}
			\centering
			\includegraphics[width=3.4in]{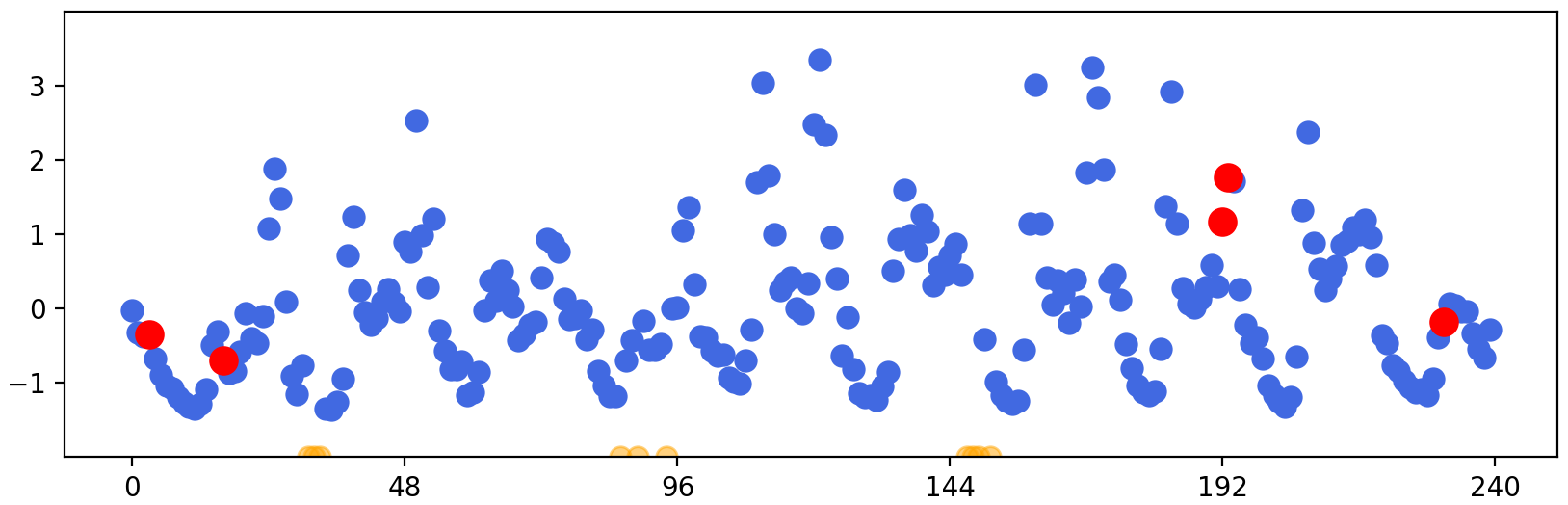}
			\label{split2}
		\end{minipage}
	}
	\\
	\subfloat[Step 1 of Stage 2]{
		\begin{minipage}[!t]{0.5\textwidth}
			\centering
			\includegraphics[width=3.4in]{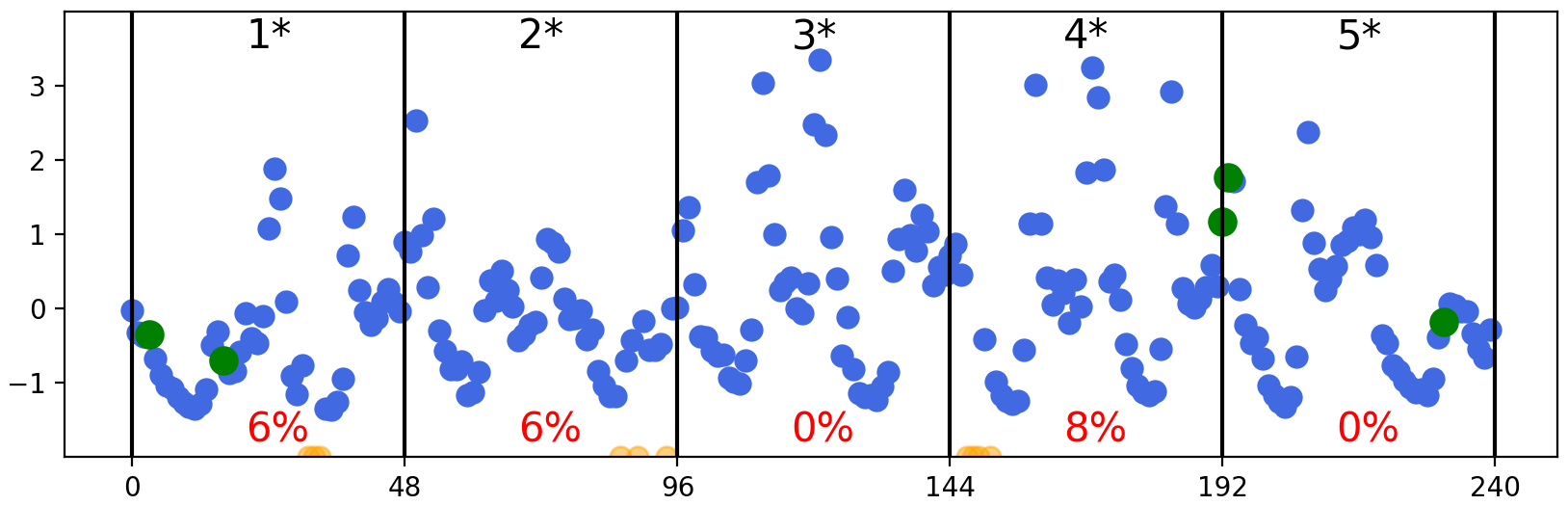}
			\label{split3}
		\end{minipage}
	}
	\\
	\subfloat[Result of Stage 2]{
		\begin{minipage}[!t]{0.5\textwidth}
			\centering
			\includegraphics[width=3.4in]{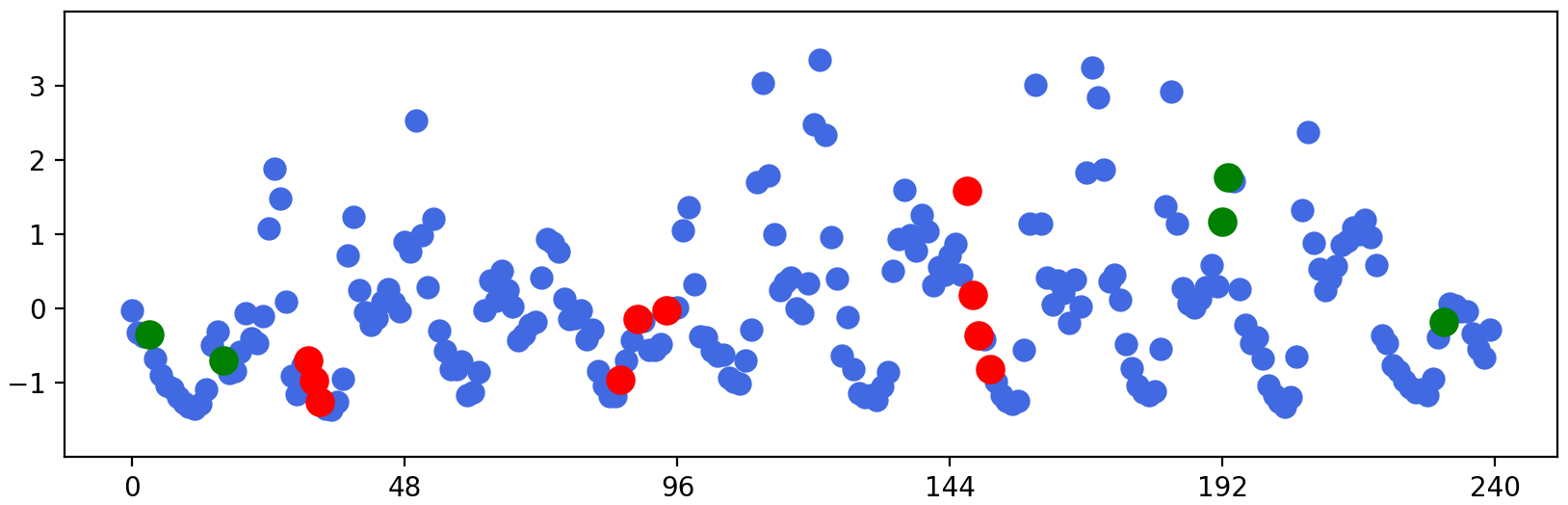}
			\label{split4}
		\end{minipage}
	}
	\\
	\caption{A toy as an example. The blue points represent the observed values, orange transparent points represent the missing values positions, red points represent the imputed values in the current stage, and green points represent the imputed values in the previous stage. The black numbers above the image represent the label for each piece of data. (*) Indicates that the data has been selected. The number of missing rates in each piece of data is indicated by the red numbers on the picture axis.}
	\label{split}
\end{figure}

An example is used to explain the entire imputation process in Fig. \ref{split}. We are given sensor data with a length of 240. First, segment the data with a length of 24 and obtain 10 pieces of data, as shown in Fig \ref{split1}, implying 10 samples. Then, filter each piece of data and bring in the data that meets the conditions to train the model. Only the second, fourth, and seventh pieces of data in this example do not meet the requirements (missing rate is less than 10\%); therefore, the remaining data (with * in Fig. \ref{split1}) are fed into the model for training. Impute the selected samples after training, as shown in Fig \ref{split2}. As the missing values are still present, we proceed to Stage 2.

In Stage 2, after receiving the imputed result from the previous stage, repeat the operation with a longer segmentation length. The segmentation result is shown in Fig \ref{split3}. After the imputation of this stage is completed, it is found that the data no longer contains missing values. Hence, we obtain the final result as shown in Fig \ref{split4}. We will only show the final result in the Experimental section. 

\section{Experimental}
The proposed framework is evaluated in this section by comparing some baselines to actual sensor data. We select NRTSI \cite{shan2021nrtsi} as the basic imputation model for the proposed framework. The results are visualized. For the dataset, we use one University of California (UCI) air quality \cite{de2008field} and four geological sensor datasets collected by physical sensors and uploaded to the GitHub \footnote{https://github.com/BomBooooo/MLSIF/tree/main}. 

\subsection{UCI Air Quality Dataset}
This dataset includes 9358 hourly averaged responses from a set of five metal oxide chemical sensors embedded in an air quality chemical multisensor device. During the experimental design process, this study employed a strategy to mitigate the reality-experimental split when artificially simulating missing values. Rather than randomly removing real data, this strategy simulates the absence of real data with a high missing rate on data with a low missing rate. Thus, the imputation result is what is required, not just better in theory. This experiment specifically selects the data with a low missing rate ( ''C6H6(GT)'', ''PT08.S1(CO)'', ''PT08.S2(NMHC)'', ''PT08.S3(NOx)'', ''PT08.S4(NO2)'', and ''PT08.S5(O3)'' ) on the dataset \cite{de2008field} and removes the data corresponding to the missing position of the data ( ''NOx(GT)'' ) with high missing rate, comparing the difference of the dropped and imputed data.

\subsubsection*{\bf Experiment 1}
First, we compare the effectiveness of the proposed multistage framework and the SIV indicator. We compared the difference between the imputation with\textbf{o}ut a multistage \textbf{f}ramework and \textbf{S}IV (OFOS), imputation with\textbf{o}ut a multistage \textbf{f}ramework but \textbf{w}ith \textbf{S}IV (OFWS), imputation \textbf{w}ith the multistage \textbf{f}ramework but with\textbf{o}ut \textbf{S}IV (WFOS), and imputation \textbf{w}ith the multistage \textbf{f}ramework and \textbf{S}IV (WFWS). 

\begin{figure*}[!t]
	\centering
	\includegraphics[width=5in]{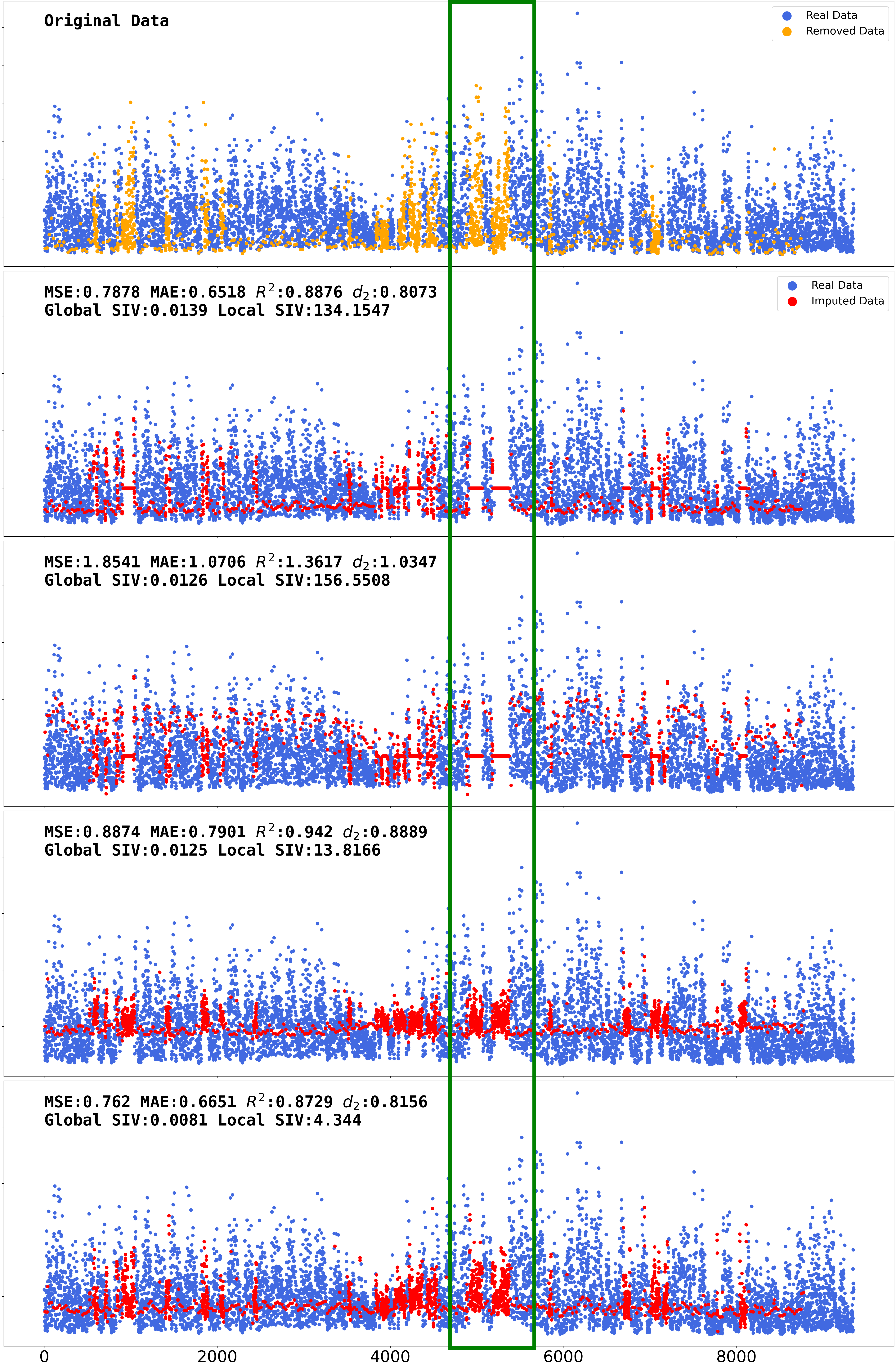}
	\caption{The result of ''C6H6(GT)'' in experiment 1. Both the second and third pictures do not use the framework; however, the loss is different. The second graph loss is MSE, and the third graph is SIV. The framework is used in the last two graphs; however, the fourth graph loss is MSE, and the last graph loss is the mixture loss. The last four pictures correspond to Case OFOS, OFWS, WFOS and WFWS.}
	\label{C6H61}
\end{figure*}

The imputation result diagram on ''C6H6(GT)'' and the metrics of all imputation results are shown in Fig. \ref{C6H61} and Table \ref{experment 1}, respectively. Other data result shown in diagrams are available on GitHub \footnote{https://github.com/BomBooooo/MLSIF/tree/main/experiment\%201}. In Fig. \ref{C6H61}, the first subpicture shows the original data, and the last four subpictures represent four imputation results of OFOS, OFWS, WFOS, and WFWS respectively.

As shown in Fig. \ref{C6H61}, the first and most obvious phenomenon is the difference between the model with and without the multistage framework. When comparing OFOS and OFWS to WFOS and WFWS, when the imputation model does not use the multistage framework, there is a clear horizontal line in the imputation result near the mean. OFOS and OFWS are rarely imputed by mean in the section of the green wireframe in Fig. \ref{C6H61}. One likely explanation is that when the missing length of the sample is similar to or equal to the length of the sample input, the model is unable to impute it due to insufficient effective information input into the model, resulting in the output being comparable to the input. This is why the fixed length model is insufficient to address the issue of long-missing segments. Consequently, the framework's benefit is clear. 

Comparing WFOS with WFWS, the model in WFOS is trained using the multistage framework and MSE loss, and the imputation results are clustered around the mean. Conversely, the imputation results of WFWS trained with the multistage framework and the mixed loss are almost consistent with the real data distribution. Even when comparing the first subimage (original data), it is difficult to distinguish the differences with unaided eyes. In terms of imputation outcomes, the WFWS appears superior to the other three models, as shown in Fig. \ref{C6H61}. 

Furthermore, we examine the quality of the imputation results based on the evaluation indicators. We use MSE, MAE, $R^2$, $d_2$, Global SIV, and Local SIV as the comparison indicators. The result is shown in Table \ref{experment 1}. Except for $R^2$ and $d_2$ indexes, which improve as their value increases, all other indexes improve as their value decreases. The best results have been highlighted in bold. Labels at the end of the values indicate better ($\uparrow$) or worse ($\downarrow$) than OFOS (baseline). 

\begin{table*}[]
	\caption{Matric Results of Experiment 1.\tnote{1}}
	\centering
	\begin{threeparttable}
		\begin{tabular}{l|c|cccccc}
			\hline
			Dataset & Case & MSE      & MAE      & $R^2$ & $d_2$     & Global SIV   & Local SIV \\
			\hline
			\multirow{4}{*}{C6H6(GT)} 
			& OFOS(Baseline) & 0.7878          & \textbf{0.6518} & 0.2074          & 0.5658          & 0.0139          & 134.1547        \\ 
			& OFWS & 1.8541 $\downarrow$ & 1.0706 $\downarrow$ & 0.0928 $\downarrow$ & 0.1423 $\downarrow$ & 0.0126 $\uparrow$ & 156.5508 $\downarrow$ \\ 
			& WFOS & 0.8874 $\downarrow$ & 0.7901 $\downarrow$ & \textbf{0.2625} $\uparrow$ & 0.5359 $\downarrow$ & 0.0125 $\uparrow$ & 13.8166 $\uparrow$ \\ 
			& WFWS & \textbf{0.7620} $\uparrow$ & 0.6651 $\downarrow$ & 0.2432 $\uparrow$ & \textbf{0.6186} $\uparrow$ & \textbf{0.0081} $\uparrow$ & \textbf{4.3440} $\uparrow$ \\ 
			\hline 
			
			\multirow{4}{*}{PT08.S1(CO)} 
			& OFOS(Baseline) & 0.7453          & 0.7095          & 0.2321          & 0.5204          & 0.1315          & 142.4815        \\ 
			& OFWS & 0.9629 $\downarrow$ & 0.8226 $\downarrow$ & 0.0205 $\downarrow$ & 0.3652 $\downarrow$ & 0.1387 $\downarrow$ & 148.3941 $\downarrow$ \\ 
			& WFOS & 0.7446 $\uparrow$ & 0.7219 $\downarrow$ & 0.2783 $\uparrow$ & 0.5471 $\uparrow$ & 0.0142 $\uparrow$ & 17.3981 $\uparrow$ \\ 
			& WFWS & \textbf{0.6839} $\uparrow$ & \textbf{0.6863} $\uparrow$ & \textbf{0.3284} $\uparrow$ & \textbf{0.6270} $\uparrow$ & \textbf{0.0108} $\uparrow$ & \textbf{12.4037} $\uparrow$ \\ 
			\hline 
			
			\multirow{4}{*}{PT08.S2(NMHC)} 
			& OFOS(Baseline) & 0.8092          & 0.7259          & 0.2459          & 0.5298          & 0.067           & 143.9391        \\ 
			& OFWS & 1.8586 $\downarrow$ & 1.0986 $\downarrow$ & 0.1162 $\downarrow$ & 0.1973 $\downarrow$ & 0.0650 $\uparrow$ & 161.8961 $\downarrow$ \\ 
			& WFOS & 0.8711 $\downarrow$ & 0.7757 $\downarrow$ & 0.2600 $\uparrow$ & 0.5052 $\downarrow$ & 0.0130 $\uparrow$ & 19.3278 $\uparrow$ \\ 
			& WFWS & \textbf{0.7369} $\uparrow$ & \textbf{0.6600} $\uparrow$ & \textbf{0.2639} $\uparrow$ & \textbf{0.6376} $\uparrow$ & \textbf{0.0125} $\uparrow$ & \textbf{9.2668} $\uparrow$ \\ 
			\hline 
			
			\multirow{4}{*}{PT08.S3(NOx)} 
			& OFOS(Baseline) & \textbf{0.5686} & \textbf{0.5782} & \textbf{0.4126} & 0.7178          & 0.0607          & 122.021         \\ 
			& OFWS & 0.9851 $\downarrow$ & 0.7594 $\downarrow$ & 0.0219 $\downarrow$ & 0.3144 $\downarrow$ & 0.0682 $\downarrow$ & 123.2433 $\downarrow$ \\ 
			& WFOS & 0.8360 $\downarrow$ & 0.6879 $\downarrow$ & 0.1234 $\downarrow$ & 0.4001 $\downarrow$ & 0.0143 $\uparrow$ & 6.3858 $\uparrow$ \\ 
			& WFWS & 0.6205 $\downarrow$ & 0.6301 $\downarrow$ & 0.3611 $\downarrow$ & \textbf{0.7477} $\uparrow$ & \textbf{0.0070} $\uparrow$ & \textbf{5.6478} $\uparrow$ \\ 
			\hline 
			
			\multirow{4}{*}{PT08.S4(NO2)} 
			& OFOS(Baseline) & 0.5817          & 0.5596          & 0.2886          & 0.5671          & 0.1045          & 131.3349        \\ 
			& OFWS & 0.7088 $\downarrow$ & 0.6466 $\downarrow$ & 0.1190 $\downarrow$ & 0.4335 $\downarrow$ & 0.1062 $\downarrow$ & 129.3343 $\uparrow$ \\ 
			& WFOS & 0.6331 $\downarrow$ & 0.5856 $\downarrow$ & 0.2395 $\downarrow$ & 0.4835 $\downarrow$ & 0.0178 $\uparrow$ & 20.3843 $\uparrow$ \\ 
			& WFWS & \textbf{0.5354} $\uparrow$ & \textbf{0.5482} $\uparrow$ & \textbf{0.3386} $\uparrow$ & \textbf{0.6609} $\uparrow$ & \textbf{0.0114} $\uparrow$ & \textbf{13.3867} $\uparrow$ \\ 
			\hline 
			
			\multirow{4}{*}{PT08.S5(O3)} 
			& OFOS(Baseline) & \textbf{0.5930} & \textbf{0.6056} & \textbf{0.2410} & \textbf{0.6032} & 0.0375          & 102.3785        \\ 
			& OFWS & 0.8888 $\downarrow$ & 0.7905 $\downarrow$ & 0.0163 $\downarrow$ & 0.3828 $\downarrow$ & 0.0426 $\downarrow$ & 111.7164 $\downarrow$ \\ 
			& WFOS & 0.6971 $\downarrow$ & 0.6867 $\downarrow$ & 0.2196 $\downarrow$ & 0.5289 $\downarrow$ & \textbf{0.0150} $\uparrow$ & 21.7660 $\uparrow$ \\ 
			& WFWS & 0.6525 $\downarrow$ & 0.6445 $\downarrow$ & 0.1418 $\downarrow$ & 0.4728 $\downarrow$ & 0.0200 $\uparrow$ & \textbf{15.6181} $\uparrow$ \\ 
			\hline 
		\end{tabular}
		\begin{tablenotes}
			\footnotesize
			\item In the comparison, the optimal index value is bolded. Labels at the end of the values indicate better ($\uparrow$) or worse ($\downarrow$) than OFOS (Baseline).
		\end{tablenotes}  
	\end{threeparttable}
	\label{experment 1}
\end{table*}

The first four indicators (MSE, MAE, $R^2$, and $d_2$) show similar trends and achieve optimal values on the same model except ''C6H6(GT)'' and ''PT08.S3(NOx)''. It can be observed that most of these indicators are optimal in WFWS, whereas some of the first four indicators are optimal in OFOS. One notable situation is that once the multistage framework is introduced, WFOS does not have a significant improvement over OFOS. One possible explanation is that after the framework is introduced, the model imputes all positions of missing data positions, whereas before the framework was introduced, the model only imputed some values, i.e., after the introduction of the framework, the model makes more attempts, and more attempts mean larger losses. After all missing values are imputed, results similar to or even better than those in OFOS can be achieved, which may help explain the effectiveness of WFWS. We believe that the absence of temporal information in the new metrics is the reason why OFWS is not better than OFOS.

The two metrics proposed in this study are shown in the last two columns. Global SIV calculates the variable of the statistical index of the overall data before and after imputation, whereas Local SIV calculates and sums the variable of the statistical index of each piece of data during the imputation process. The introduction of the multistage framework considerably improves the imputation results on these two metrics. Among them, the most noticeable improvement is in Local SIV, and the value is reduced by dozens. More details about the relationship between the specific trend of these two indicators and other indicators as well as the relationship between these two indicators and the imputation results are further explored in Experiment 2. 

From Experiment 1, it is shown that the use of the framework can solve the issue of insufficient imputation caused by fixed length and one-stage. Furthermore, using the mixed loss can improve the imputation effect of the model and alleviate the problem of MSE as the model loss. 

\subsubsection*{\bf Experiment 2}
\begin{figure*}[htbp]
	\centering
	\subfloat[$\alpha=0.00$]{
		\begin{minipage}[!t]{0.3\textwidth}
			\centering
			\includegraphics[width=1\textwidth]{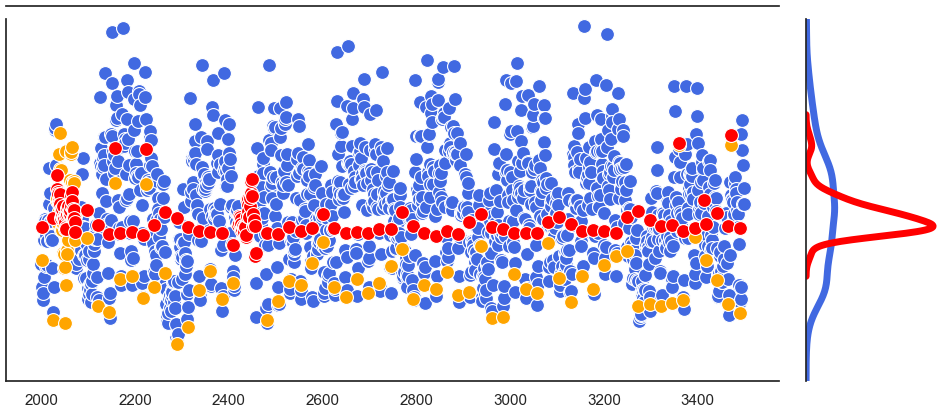}
		\end{minipage}
	}
	\subfloat[$\alpha=0.25$]{
		\begin{minipage}[!t]{0.3\textwidth}
			\centering
			\includegraphics[width=1\textwidth]{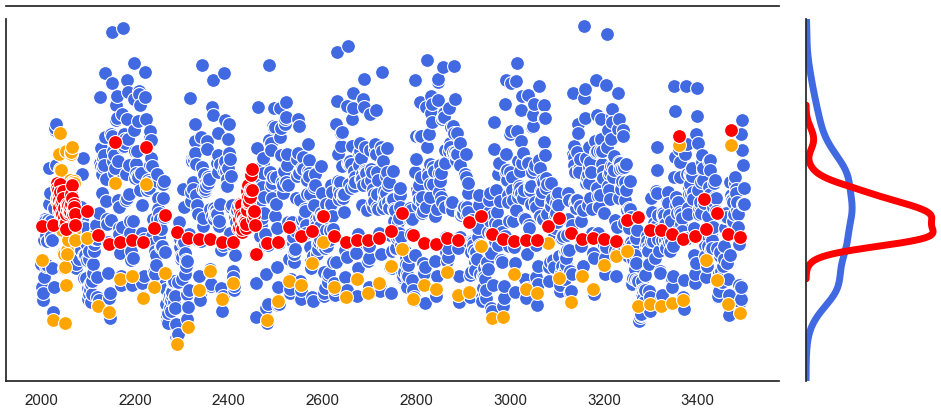}
		\end{minipage}
	}
	\subfloat[$\alpha=0.50$]{
		\begin{minipage}[!t]{0.3\textwidth}
			\centering
			\includegraphics[width=1\textwidth]{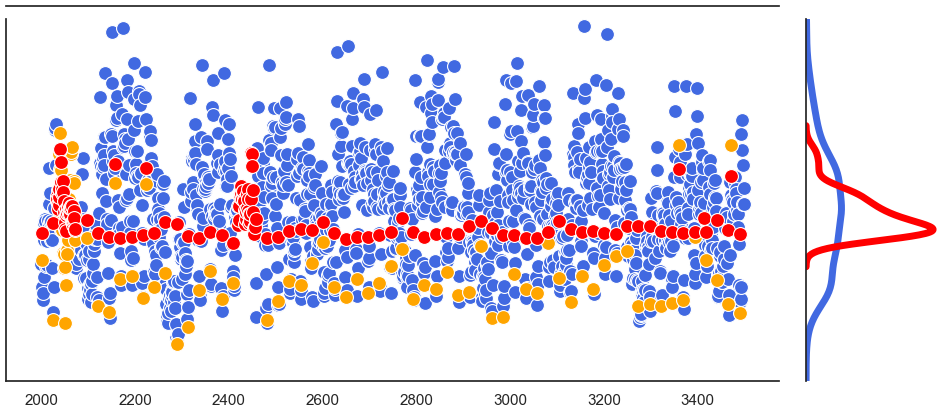}
		\end{minipage}
	}
	\\ 
	\subfloat[$\alpha=0.75$]{
		\begin{minipage}[!t]{0.3\textwidth}
			\centering
			\includegraphics[width=1\textwidth]{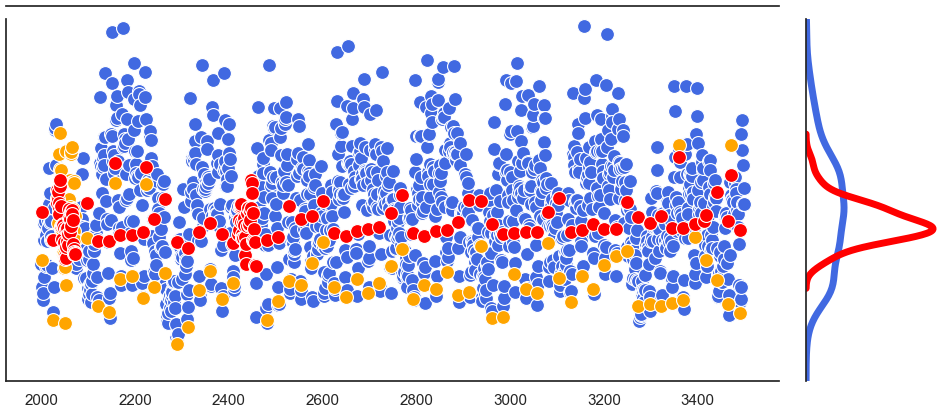}
		\end{minipage}
	}
	\subfloat[$\alpha=0.80$]{
		\begin{minipage}[!t]{0.3\textwidth}
			\centering
			\includegraphics[width=1\textwidth]{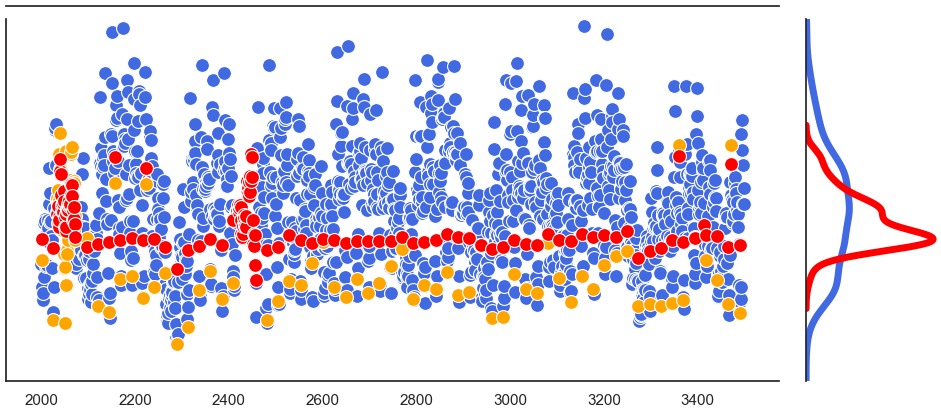}
		\end{minipage}
	}
	\subfloat[$\alpha=0.90$]{
		\begin{minipage}[!t]{0.3\textwidth}
			\centering
			\includegraphics[width=1\textwidth]{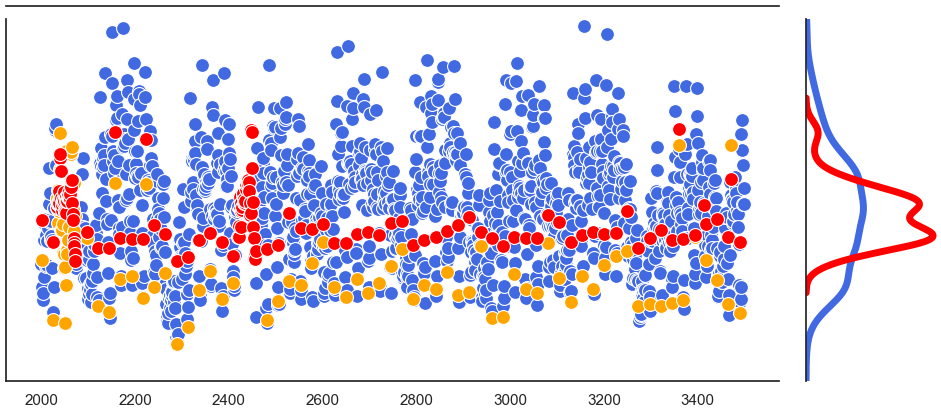}
		\end{minipage}
	}
	\\ 
	\subfloat[$\alpha=0.95$]{
		\begin{minipage}[!t]{0.3\textwidth}
			\centering
			\includegraphics[width=1\textwidth]{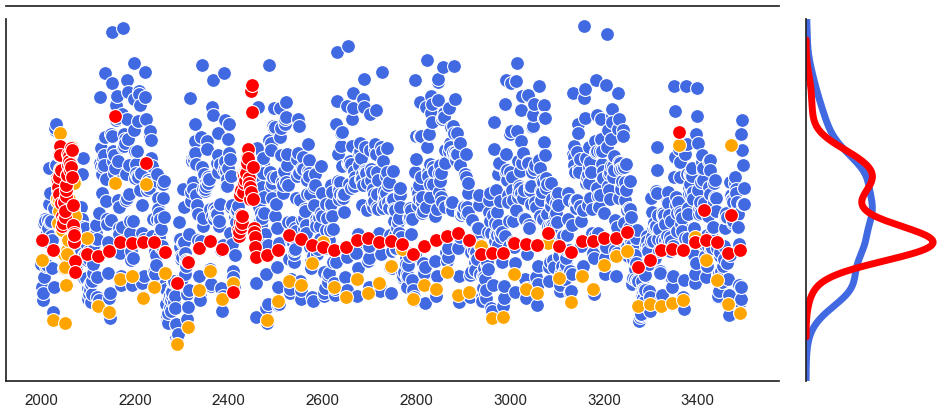}
		\end{minipage}
	}
	\subfloat[$\alpha=0.96$]{
		\begin{minipage}[!t]{0.3\textwidth}
			\centering
			\includegraphics[width=1\textwidth]{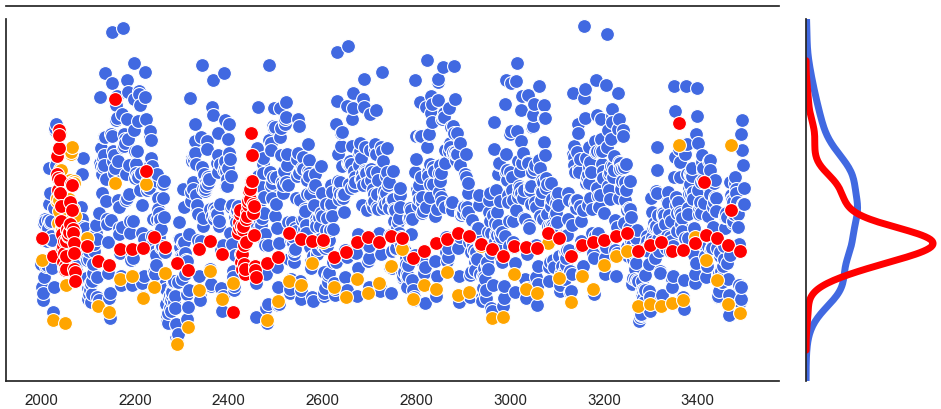}
		\end{minipage}
	}
	\subfloat[$\alpha=0.97$]{
		\begin{minipage}[!t]{0.3\textwidth}
			\centering
			\includegraphics[width=1\textwidth]{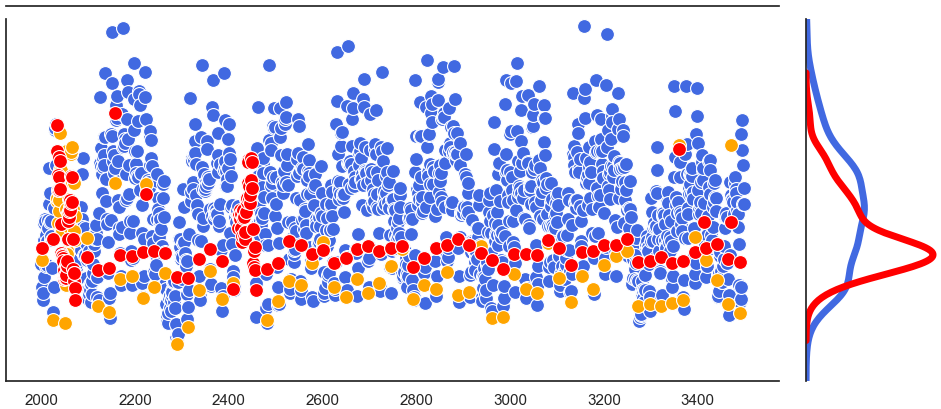}
		\end{minipage}
	}
	\\ 
	\subfloat[$\alpha=0.98$]{
		\begin{minipage}[!t]{0.3\textwidth}
			\centering
			\includegraphics[width=1\textwidth]{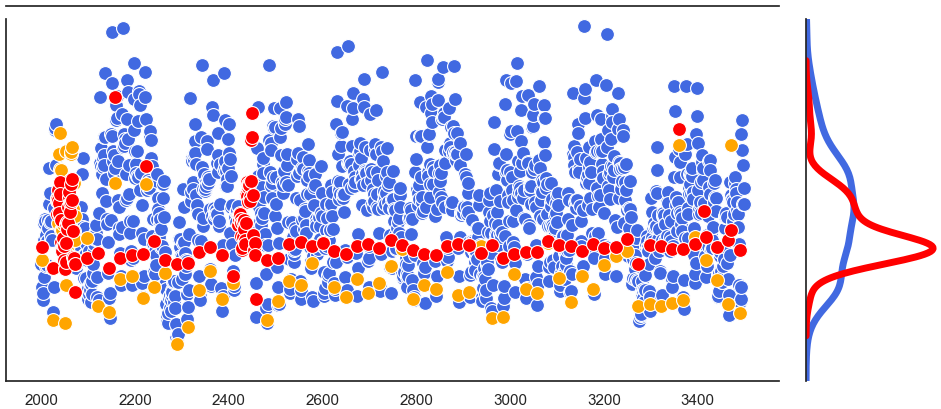}
		\end{minipage}
	}
	\subfloat[$\alpha=0.99$]{
		\begin{minipage}[!t]{0.3\textwidth}
			\centering
			\includegraphics[width=1\textwidth]{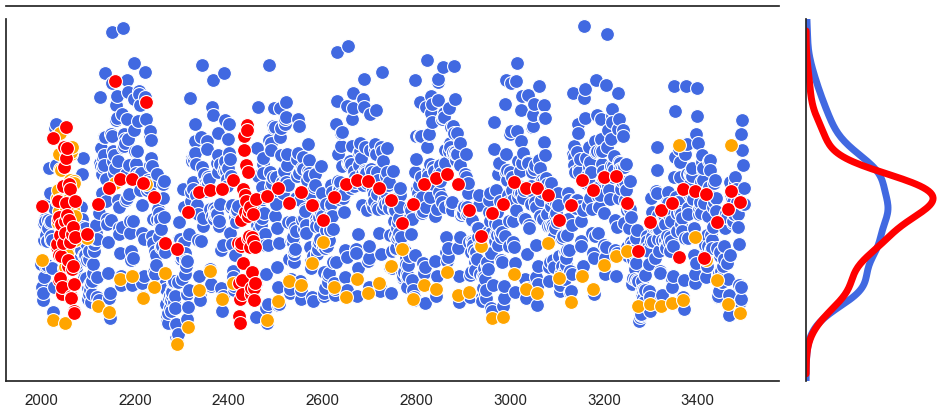}
		\end{minipage}
	}
	\subfloat[$\alpha=1.00$]{
		\begin{minipage}[!t]{0.3\textwidth}
			\centering
			\includegraphics[width=1\textwidth]{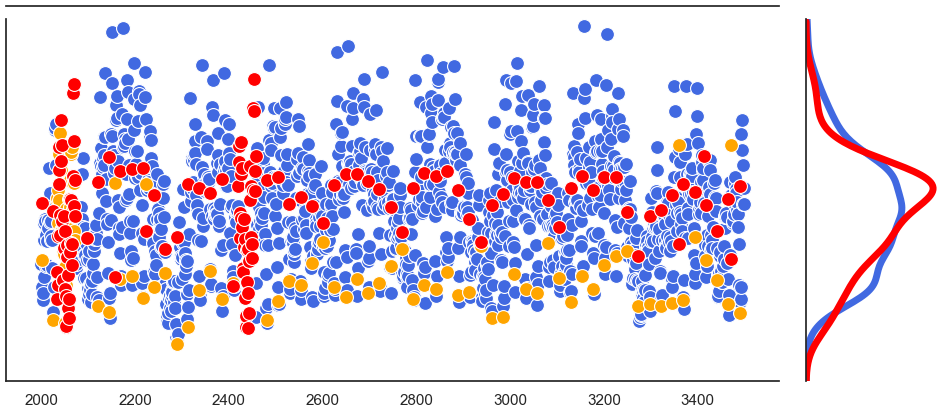}
		\end{minipage}
	}
	
	\caption{Imputation results for small missing gaps in data ''PT08.S2(NMHC)'', with the blue points representing observation values, orange points representing removed values, and red points representing imputed values. The blue and red lines on the right side of each image represent the distribution of the original and imputed data, respectively.}
	\label{212}
\end{figure*}

\begin{figure*}[htbp]
	\centering
	\subfloat[$\alpha=0.00$]{
		\begin{minipage}[!t]{0.3\textwidth}
			\centering
			\includegraphics[width=1\textwidth]{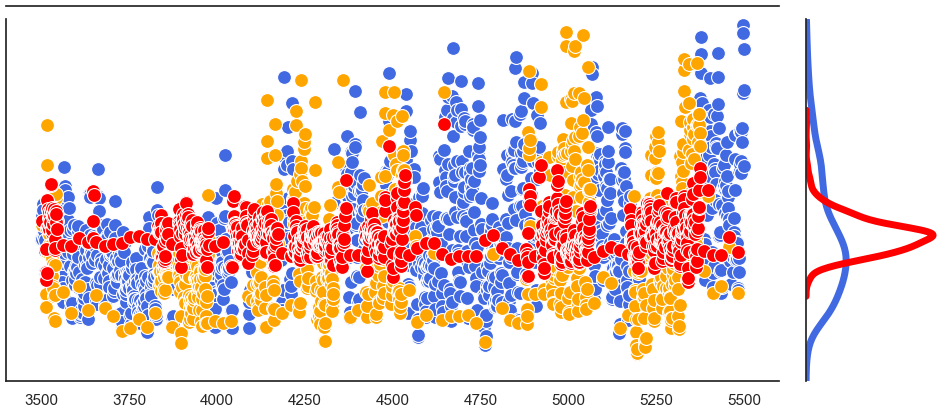}
		\end{minipage}
	}
	\subfloat[$\alpha=0.25$]{
		\begin{minipage}[!t]{0.3\textwidth}
			\centering
			\includegraphics[width=1\textwidth]{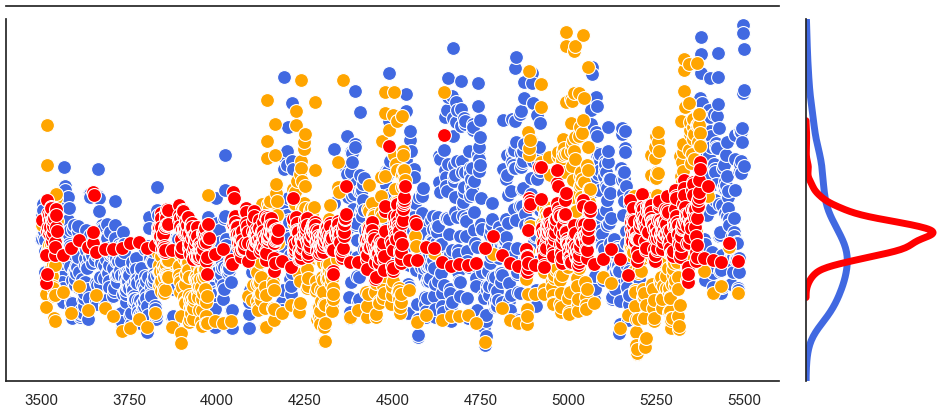}
		\end{minipage}
	}
	\subfloat[$\alpha=0.50$]{
		\begin{minipage}[!t]{0.3\textwidth}
			\centering
			\includegraphics[width=1\textwidth]{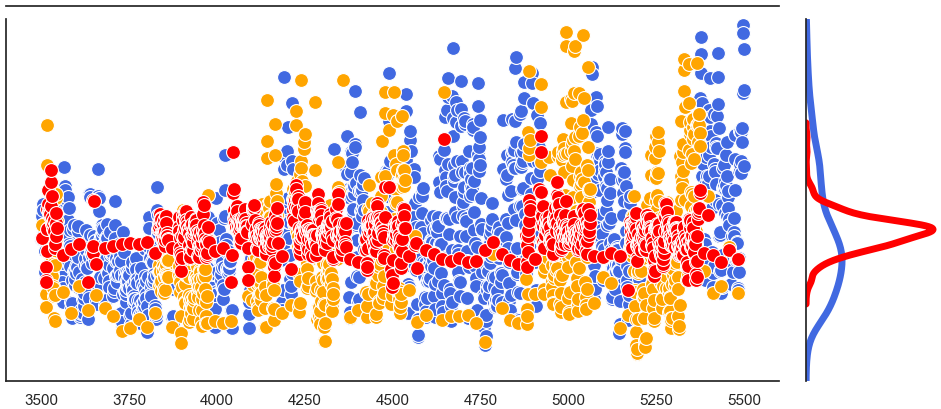}
		\end{minipage}
	}
	\\ 
	\subfloat[$\alpha=0.75$]{
		\begin{minipage}[!t]{0.3\textwidth}
			\centering
			\includegraphics[width=1\textwidth]{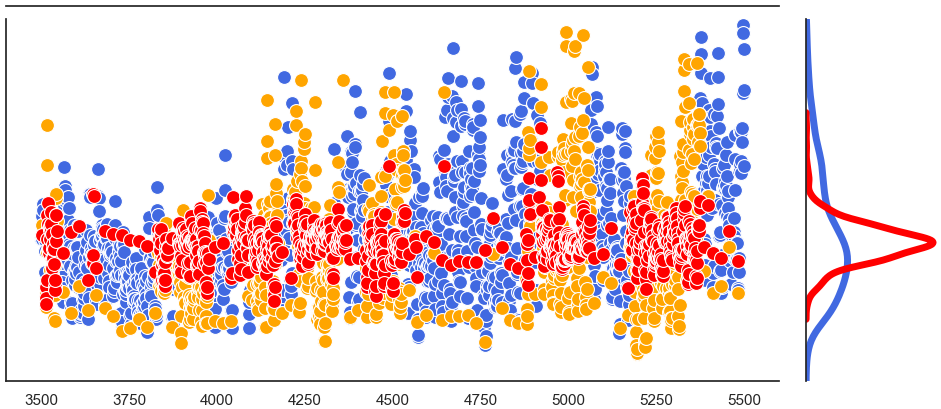}
		\end{minipage}
	}
	\subfloat[$\alpha=0.80$]{
		\begin{minipage}[!t]{0.3\textwidth}
			\centering
			\includegraphics[width=1\textwidth]{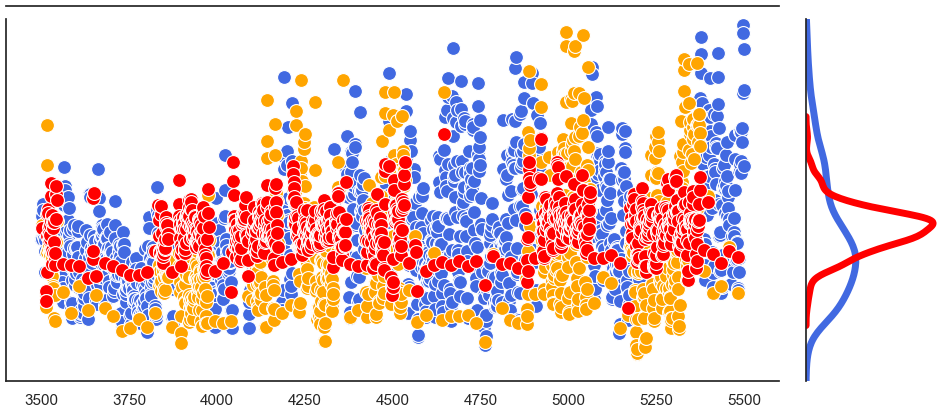}
		\end{minipage}
	}
	\subfloat[$\alpha=0.90$]{
		\begin{minipage}[!t]{0.3\textwidth}
			\centering
			\includegraphics[width=1\textwidth]{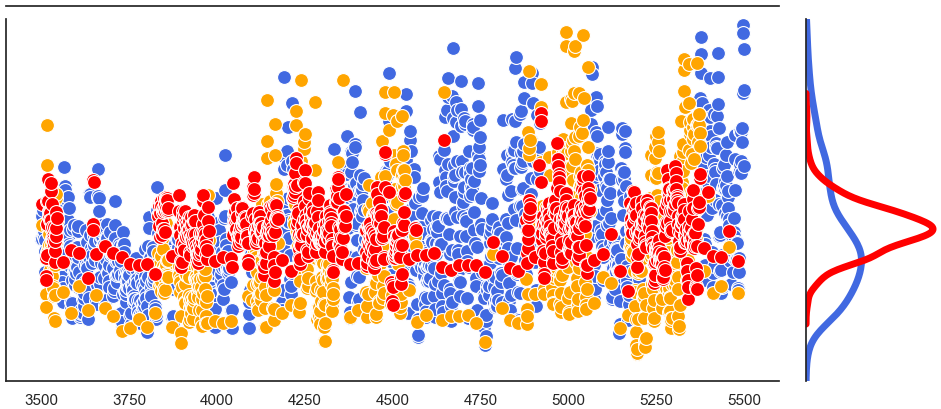}
		\end{minipage}
	}
	\\ 
	\subfloat[$\alpha=0.95$]{
		\begin{minipage}[!t]{0.3\textwidth}
			\centering
			\includegraphics[width=1\textwidth]{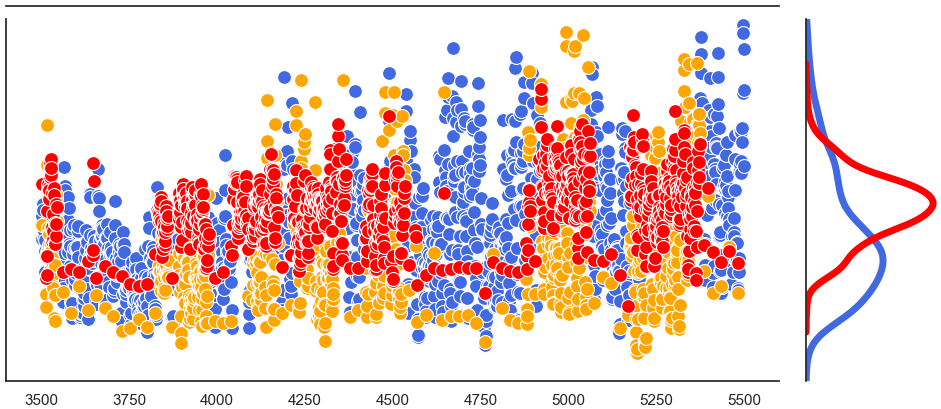}
		\end{minipage}
	}
	\subfloat[$\alpha=0.96$]{
		\begin{minipage}[!t]{0.3\textwidth}
			\centering
			\includegraphics[width=1\textwidth]{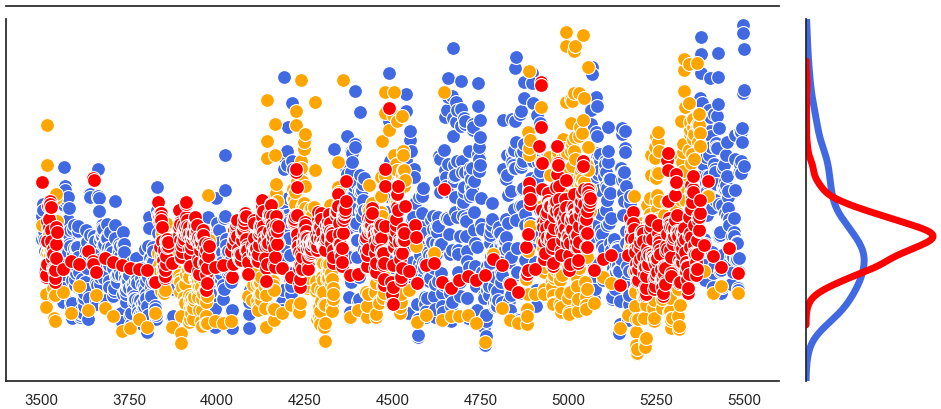}
		\end{minipage}
	}
	\subfloat[$\alpha=0.97$]{
		\begin{minipage}[!t]{0.3\textwidth}
			\centering
			\includegraphics[width=1\textwidth]{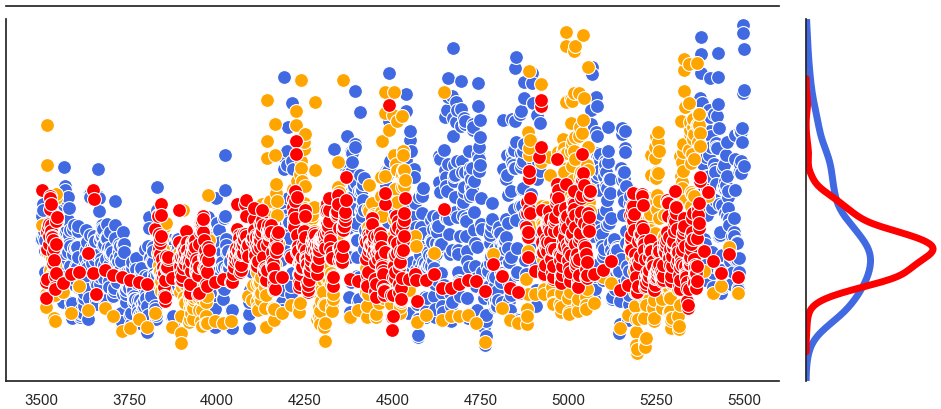}
		\end{minipage}
	}
	\\ 
	\subfloat[$\alpha=0.98$]{
		\begin{minipage}[!t]{0.3\textwidth}
			\centering
			\includegraphics[width=1\textwidth]{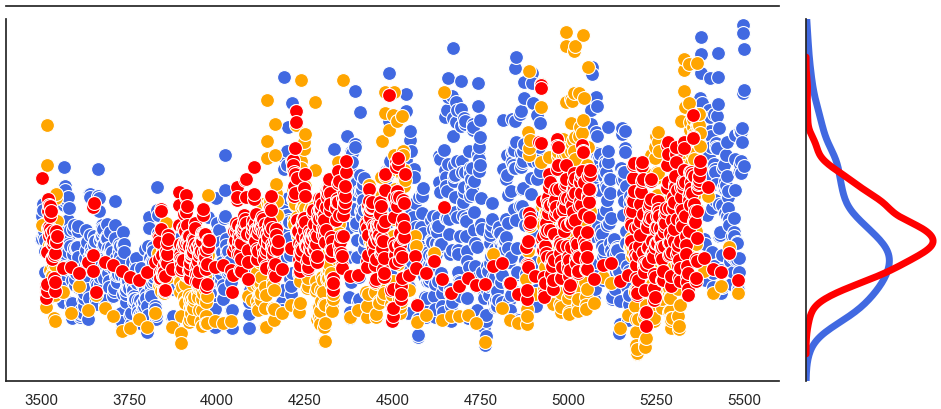}
		\end{minipage}
	}
	\subfloat[$\alpha=0.99$]{
		\begin{minipage}[!t]{0.3\textwidth}
			\centering
			\includegraphics[width=1\textwidth]{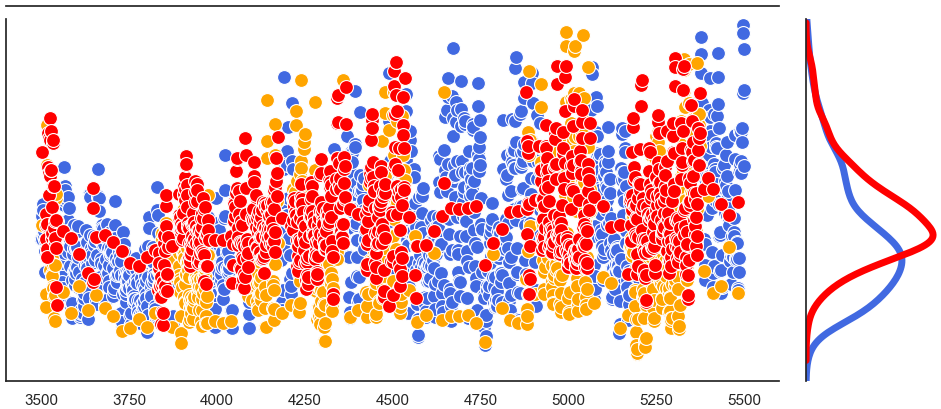}
		\end{minipage}
	}
	\subfloat[$\alpha=1.00$]{
		\begin{minipage}[!t]{0.3\textwidth}
			\centering
			\includegraphics[width=1\textwidth]{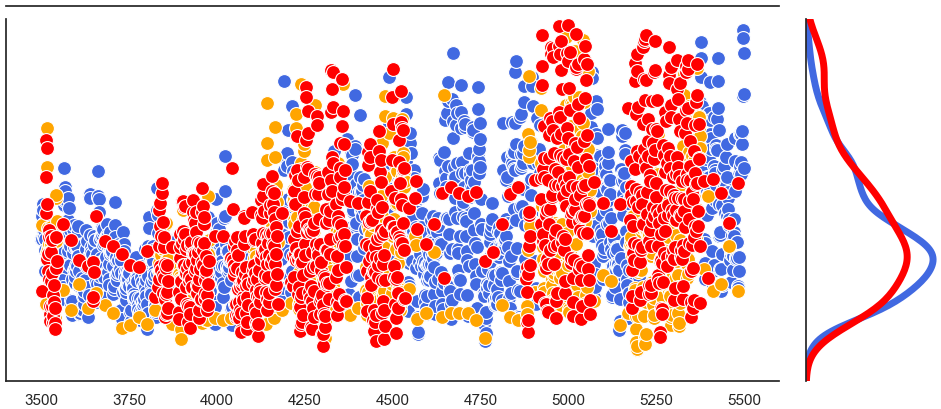}
		\end{minipage}
	}
	
	\caption{Imputation results for large missing gaps in data ''PT08.S2(NMHC)'', with the blue points representing observation values, orange points representing removed values, and red points representing imputed values. The blue and red lines on the right side of each image represent the distribution of the original and imputed data, respectively.}
	\label{222}
\end{figure*}

\begin{figure*}[!t]
	\centering
	\subfloat[MSE]{
		\begin{minipage}[!t]{0.45\textwidth}
			\centering
			\includegraphics[width=1\textwidth]{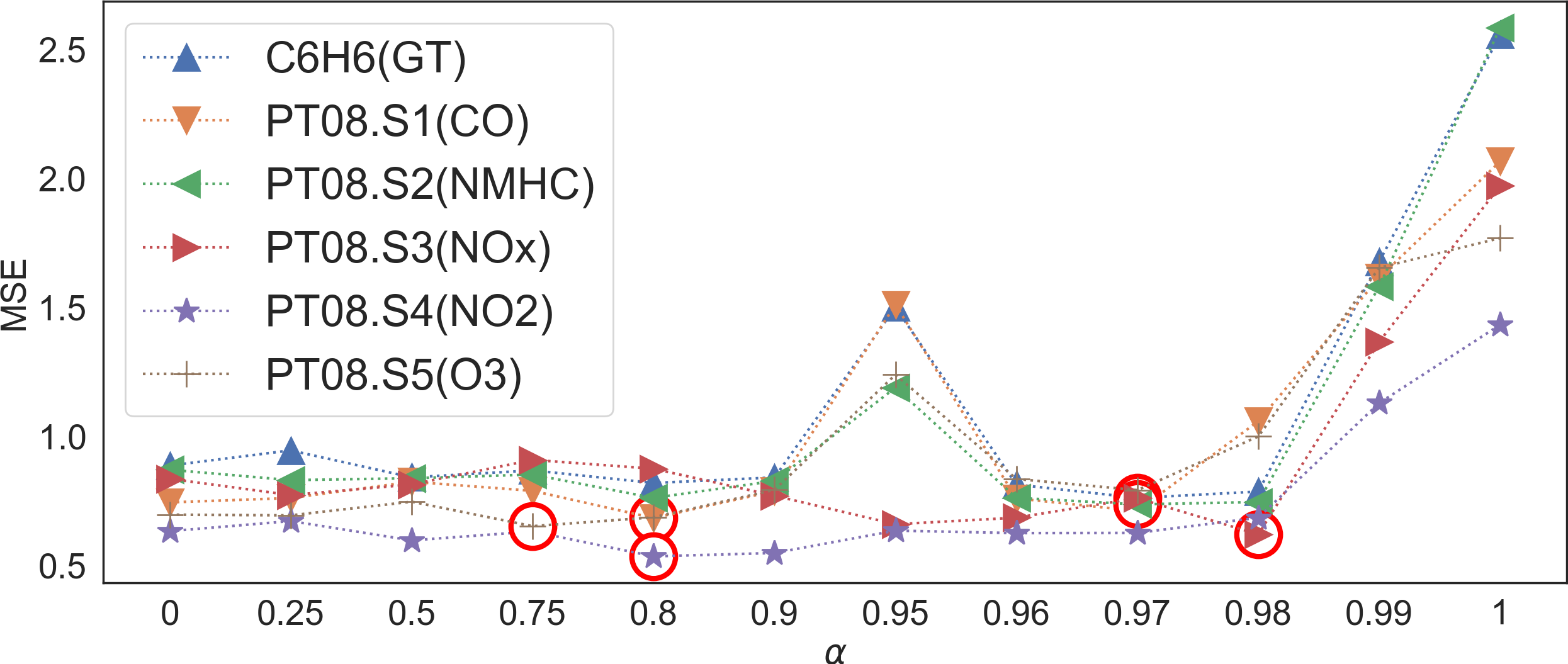}
		\end{minipage}
	\label{mse matric}
	}
	\subfloat[MAE]{
		\begin{minipage}[!t]{0.45\textwidth}
			\centering
			\includegraphics[width=1\textwidth]{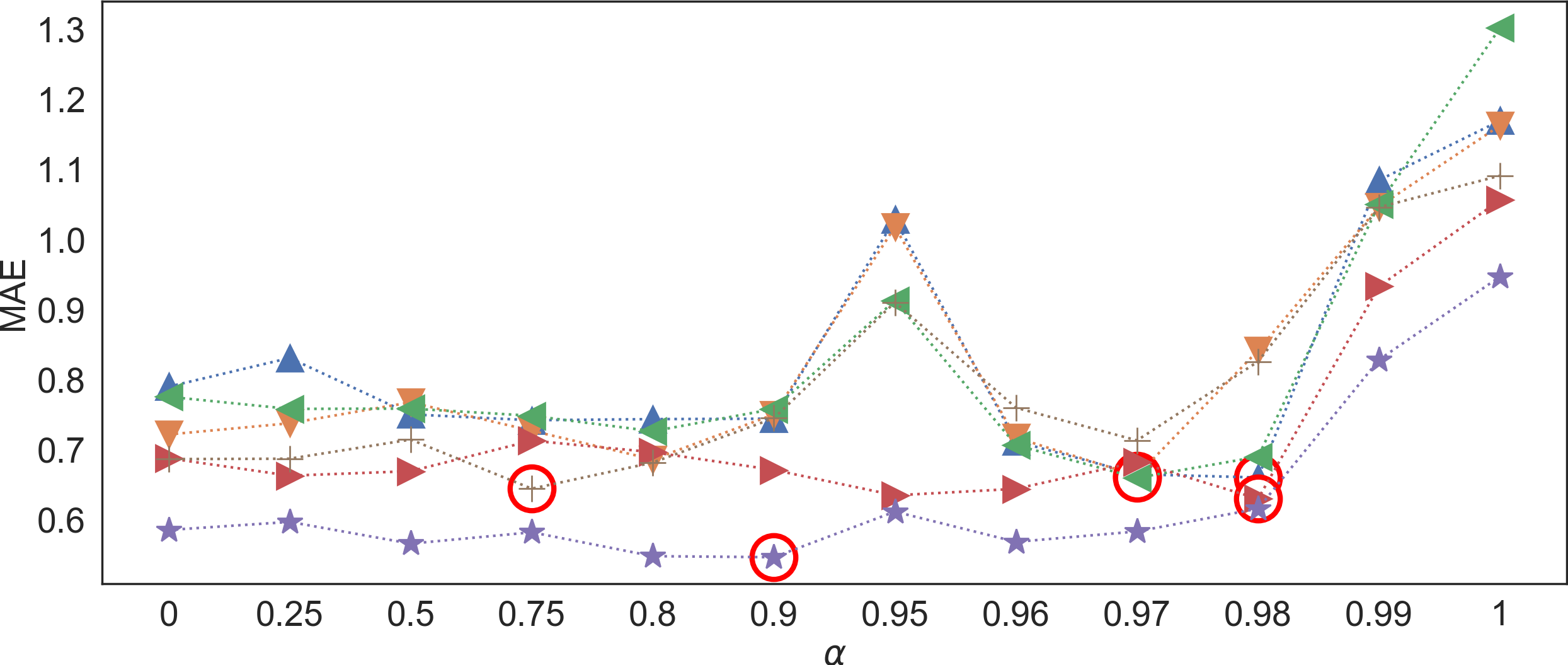}
		\end{minipage}
	}
	\\ 
	\subfloat[$R^2$]{
		\begin{minipage}[!t]{0.45\textwidth}
			\centering
			\includegraphics[width=1\textwidth]{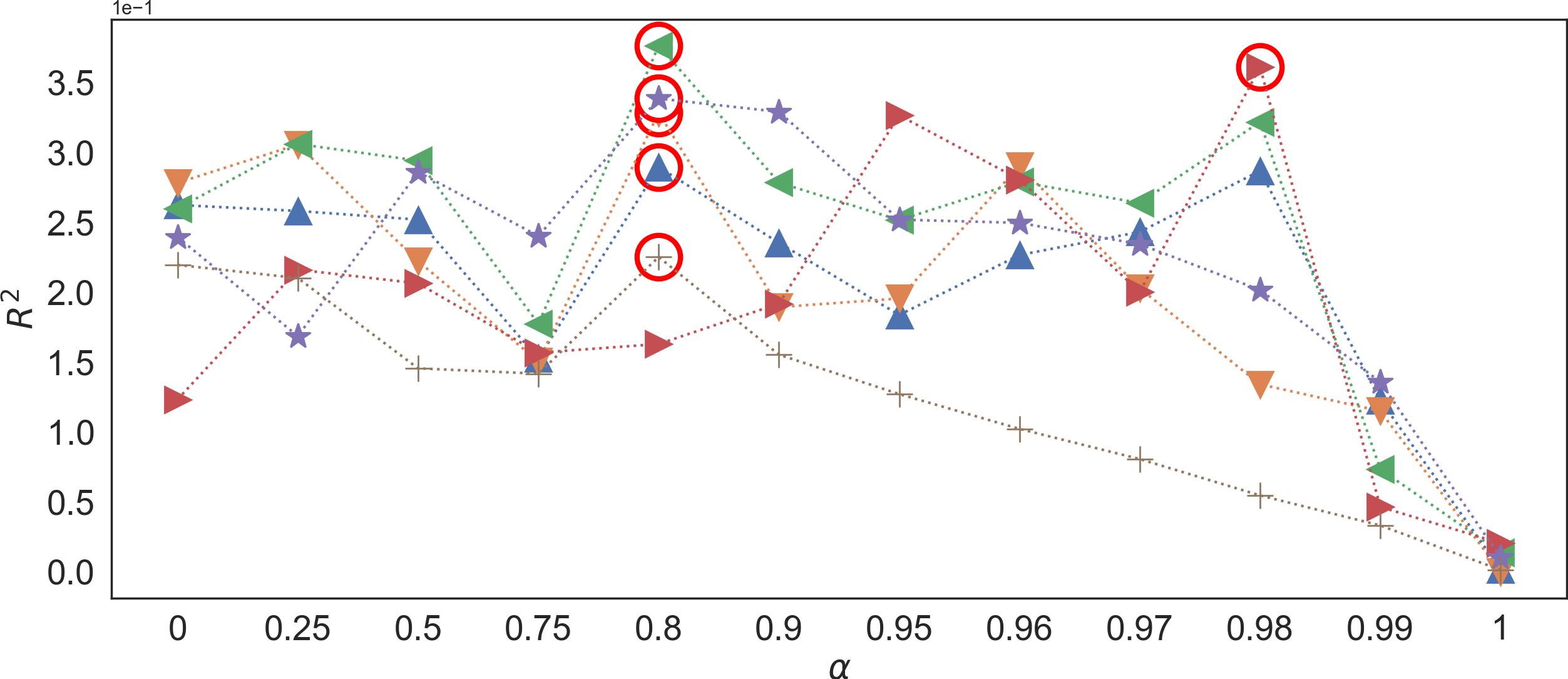}
		\end{minipage}
	}
	\subfloat[$d_2$]{
		\begin{minipage}[!t]{0.45\textwidth}
			\centering
			\includegraphics[width=1\textwidth]{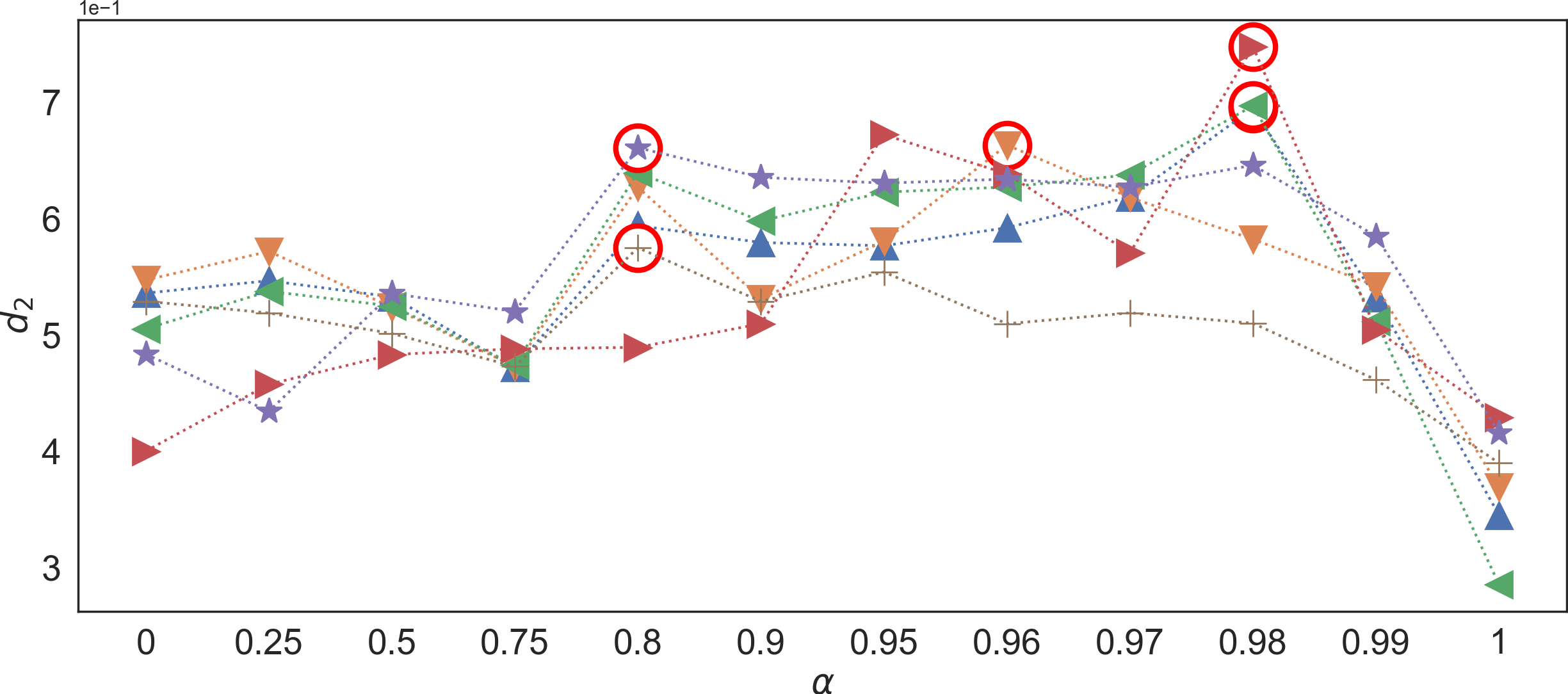}
		\end{minipage}
	\label{d_2 matric}
	}
	\\ 
	\subfloat[Global SIV]{
		\begin{minipage}[!t]{0.45\textwidth}
			\centering
			\includegraphics[width=1\textwidth]{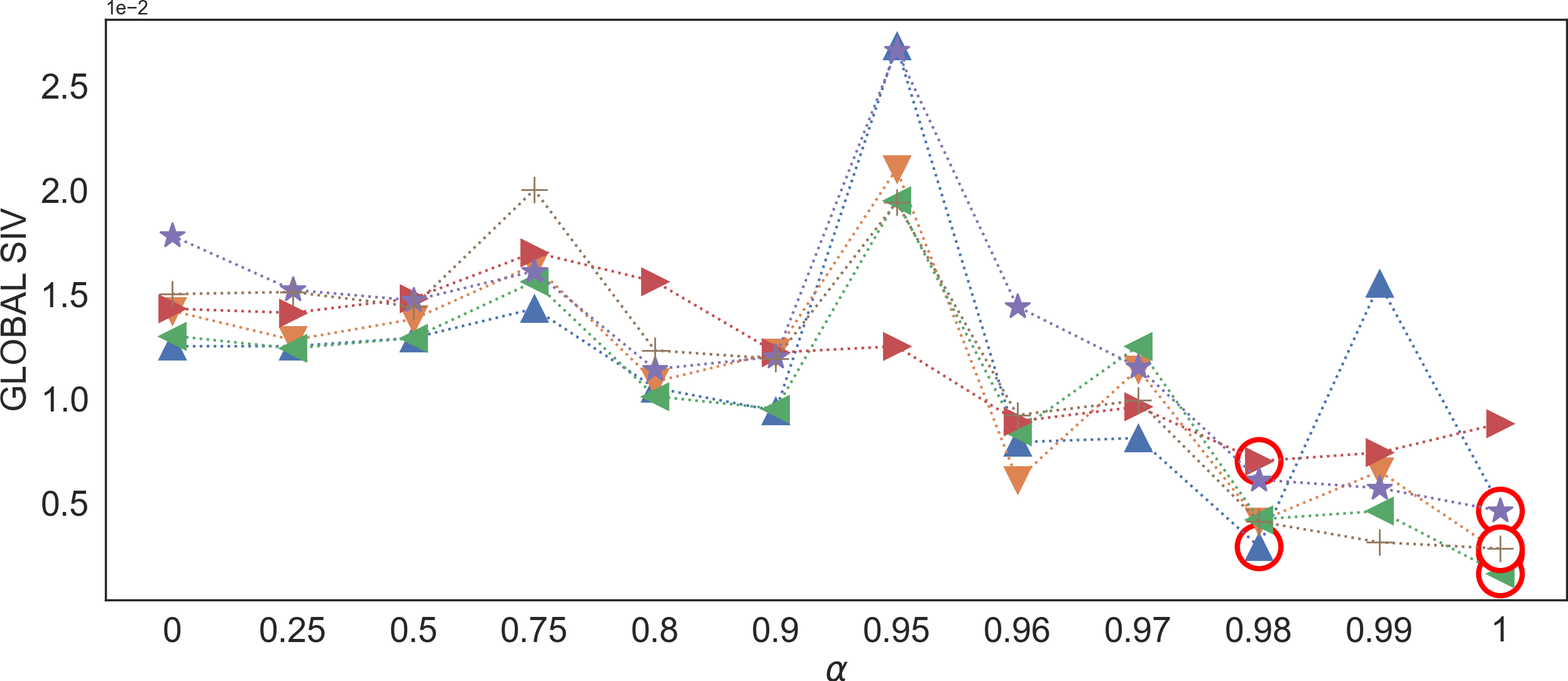}
		\end{minipage}
	\label{gsiv}
	}
	\subfloat[Local SIV]{
		\begin{minipage}[!t]{0.45\textwidth}
			\centering
			\includegraphics[width=1\textwidth]{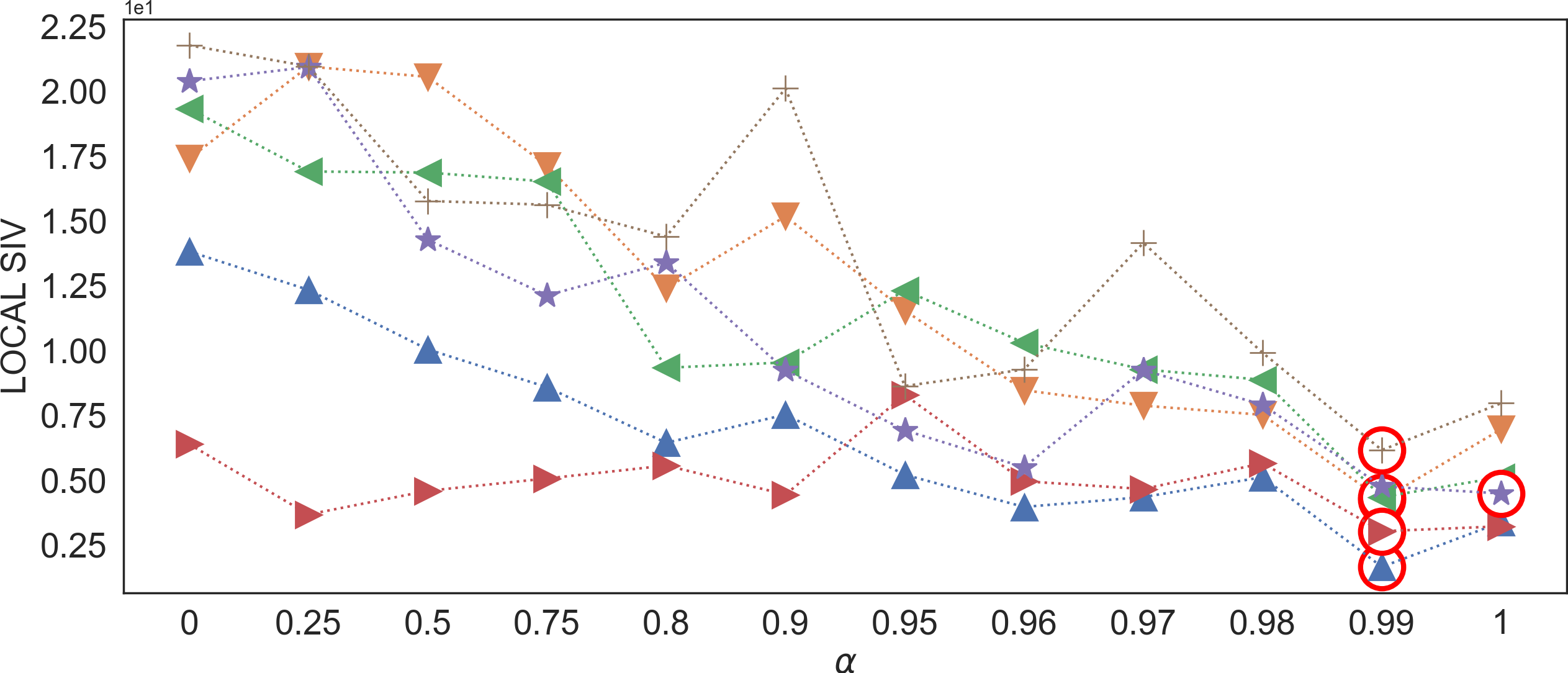}
		\end{minipage}
	\label{lsiv}
	}
	\\ 
	\caption{Metrics change with $\alpha$. The abscissa represents the value of $\alpha$, and the ordinate represents the value of each indicator.}
	\label{matrics}
\end{figure*}

In this experiment, we explore the impact of the weights $\alpha$ between MSE and SIV in the mixed loss. Simultaneously, the relationship between SIV and MSE as a metric is determined by analyzing the imputation results. Here, we present and analyze the imputation results of ''PT08.S2(NMHC)'', as shown in Fig. \ref{212} and Fig. \ref{222}. They show the model's imputation details for small- and large-missing gaps. Other compared results on other datasets can be found on the GitHub \footnote{https://github.com/BomBooooo/MLSIF/tree/main/experiment\%202}. In Fig. \ref{212} and Fig. \ref{222}, the blue points represent the real data, red points represent the imputed data, and orange points represent the removed data. The distribution of the original and imputed data is represented by the blue and red lines on the right side of each image, respectively. The corresponding variation law of each index with $\alpha$ is explained and shown in Fig. \ref{matrics}, where the dots circled in red represent the optimal value's location. 

In Fig. \ref{212}, the presence of a horizontal line close to the mean throughout the imputation results is the most noticeable. This phenomenon is alleviated when the $\alpha$ is greater than 0.98. The same phenomenon can also be observed from the data distribution diagram. When $\alpha$ is low, the imputed data cluster around the mean and gradually spreads out as $\alpha$ increases. When $\alpha=1$, the distribution of imputed values is almost identical to that distribution of the original data. 

In Fig. \ref{222}, there are large missing gaps in the data. The most obvious phenomenon from the distribution map is that the imputed data become looser as $\alpha$ increases. When $\alpha$ is low, the imputed data tend to cluster around the mean. As $\alpha$ increases, the data becomes more dispersed, and, when $\alpha$ equals 1, the data is completely dispersed, with the degree of dispersion being close to that of the original data. For the above two losses, MSE-type loss can capture certain temporal characteristics; however, it is prone to making imputations around the mean. SIV cannot capture temporal characteristics but allows the model to learn the data's discrete situation; therefore, combining the two yields better results. Figures \ref{212} and \ref{222} show that most models perform well with small
missing gaps. However, MLSIF impressively solves large-missing gap problems. 

In Fig. \ref{mse matric} - \ref{d_2 matric}, the changing trend of these indicators is consistent. Generally, as $\alpha$ increases, the metrics improve first and then deteriorate. When $\alpha$ is small, the changes in these indicators are subtle, and most optimal values are found between 0.8 and 0.98. When $\alpha$ exceeds 0.98, the indicators rapidly deteriorate. The reason for this is that when $\alpha$ is large, MSE plays an insignificant role in the loss, allowing the SIV to play a dominant role, with the imputed data scattered within a certain range, resulting in these indicators having large values. One thing is certain: using the SIV to form the mixed loss improves the results on both metrics; however, using only the SIV makes the result worse than using only the MSE. 

In Fig. \ref{gsiv} and \ref{lsiv}, the Global SIV and Local SIV metrics show a clear downward trend. This result was consistent with expectations. As the proportion of SIV loss increases, the value of the final imputed result using SIV as a metric also decreases. Consequently, unlike the others, these two indicators did not change significantly at $\alpha=0.98$. The optimal values are all obtained at around $\alpha=0.99$. At this time, the SIV plays a dominant role in the loss. It shows that between $\alpha=0.98$ and $\alpha=0.99$, MSE and SIV can be traded off, resulting in both being relatively low. 

By observing and analyzing the experimental results, we discovered that the Global SIV is more concerned with the difference between the imputed and original value. The Global SIV is not particularly large if there are few points that deviate from the discrete degree of the original data. The Local SIV calculates the changes in four statistical indicators for each small sample. Consequently, when the Global SIV is smaller, the data appear more compact. Furthermore, when the Local SIV is smaller, the data are more coherent. Therefore, the Local SIV can be used to assess the consistency between the imputed and true value. 

In summary, Experiment 2 investigates the mixing loss further and investigated its impact on various indicators and imputation results. The experimental results show that using the mixed loss is better than using the MSE or the SIV loss alone in all metrics. Furthermore, it was found that most of the optimal values are obtained at around $\alpha=0.98$. These six measurement methods correspond to three types of measurement angles as metrics. The first type (MSE, MAE) focuses on the difference between each imputation and removed points. The second type ($R^2, d_2$) focuses on the distribution between the imputed and mean of the dropped data. The third type (SIV) focuses on the distribution difference between the original and imputed data.

\subsection{Physical Sensor Data}
Due to the difficulty of acquiring and maintaining geological sensor data, there are frequently a significant number of missing values, making the subsequent analysis and research difficult. The sensitive information of the timestamp and numerical value of these data have been processed and uploaded to the GitHub \footnote{https://github.com/BomBooooo/MLSIF/tree/main/}.

\subsubsection*{\bf Experiment 3}
The selected data in this experiment, named 35717443\_temp, contains more than 20,000 data points and has approximately 3\% missing values. In this section, we test the framework's performance on these data with different missing rate scenarios. In contrast to other methods for randomly constructing missing values, the strategy used in this study is to construct large segments of missing values, i.e., a random position is selected and a portion of the original data is removed near this position. We compare the following methods when confronted with such a dataset:

\begin{itemize}
	\item{ZOO (From R Package) \cite{zeileis2005zoo}: three of the methods were chosen, namely, mean imputation, interpolation imputation and (cubic) linear imputation, called na.aggregate, na.approx, and na.spline, respectively.}
	\item{ImputeTS (From R Package) \cite{moritz2017imputets}: we choose two of these methods, namely, linear imputation and structural model and Kalman smooth imputation, named na.interpolation and na.kalman, respectively.}
	\item{BRITS and BRITS-I \cite{cao2018brits}: a method based on recurrent neural networks
		for missing value imputation in time-series data.}
	\item{CSDI \cite{tashiro2021csdi}: a time-series imputation method that utilizes score-based diffusion
		models conditioned on observed data.}
	\item{NAOMI \cite{liu2019naomi}: a nonautoregressive approach to impute long-range sequences given arbitrary missing patterns.}
	\item{NRTSI \cite{shan2021nrtsi}: reformulate time series as permutation-equivariant sets do not impose any recurrent structures to impute missing data.}
\end{itemize}

\begin{figure*}[!t]
	\centering
	\subfloat[0\% missing]{
		\begin{minipage}[!t]{0.45\textwidth}
			\centering
			\includegraphics[width=1\textwidth]{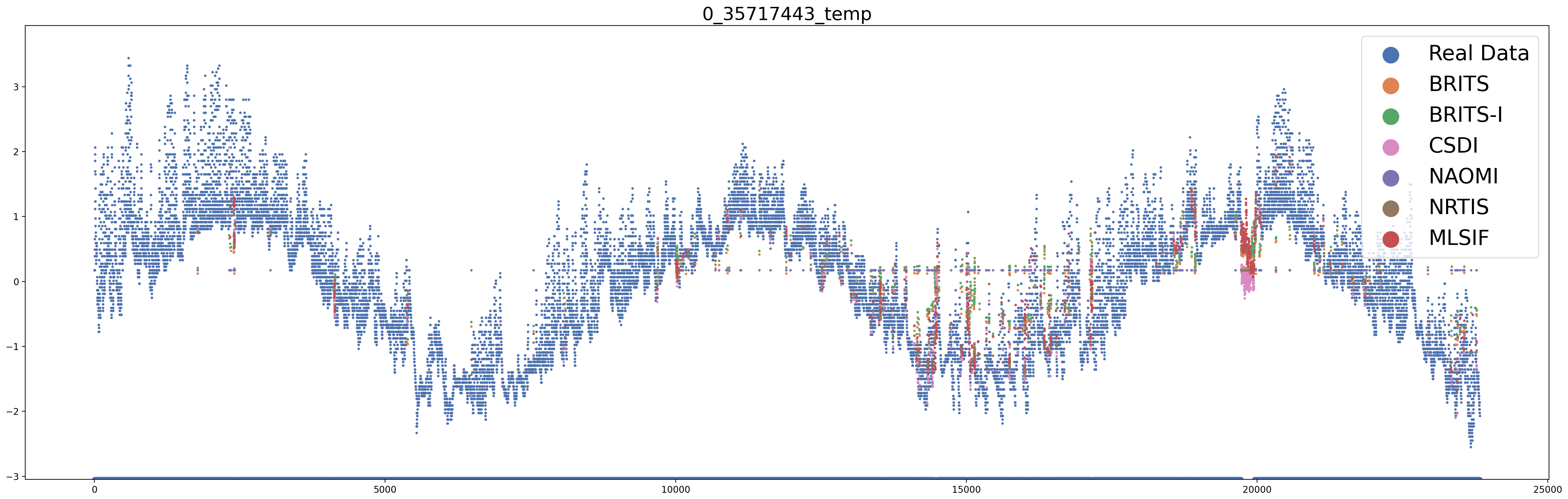}
		\end{minipage}
		\label{0 missing}
	}
	\subfloat[10\% missing]{
		\begin{minipage}[!t]{0.45\textwidth}
			\centering
			\includegraphics[width=1\textwidth]{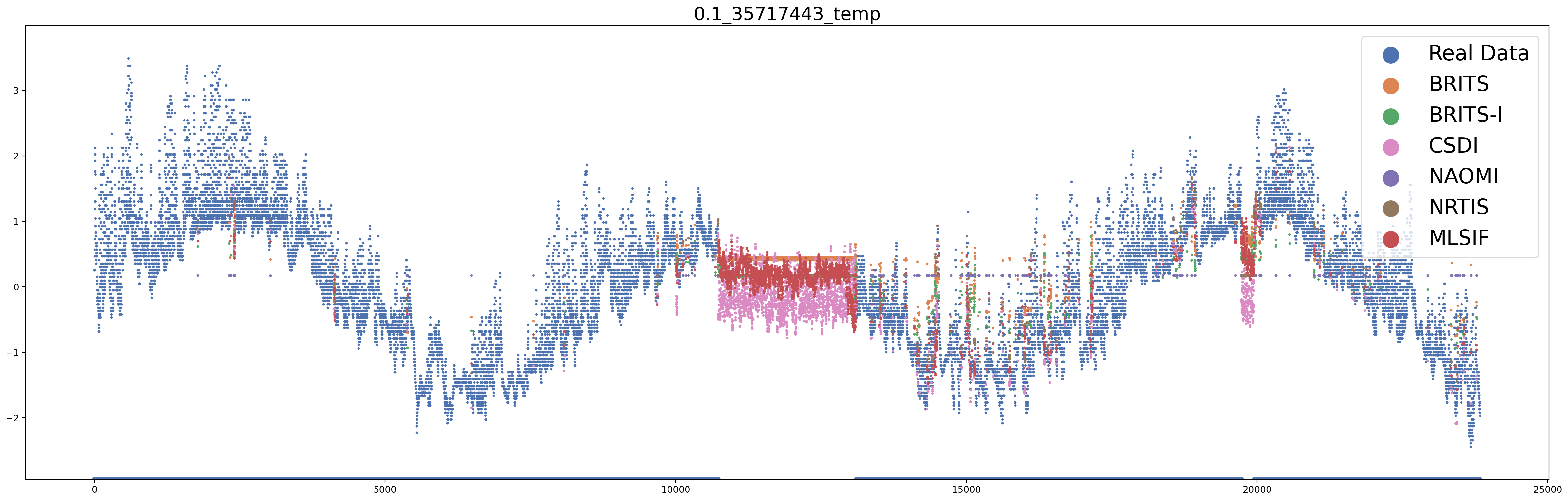}
		\end{minipage}
		\label{10 missing}
	}
	\\ 
	\subfloat[20\% missing]{
		\begin{minipage}[!t]{0.45\textwidth}
			\centering
			\includegraphics[width=1\textwidth]{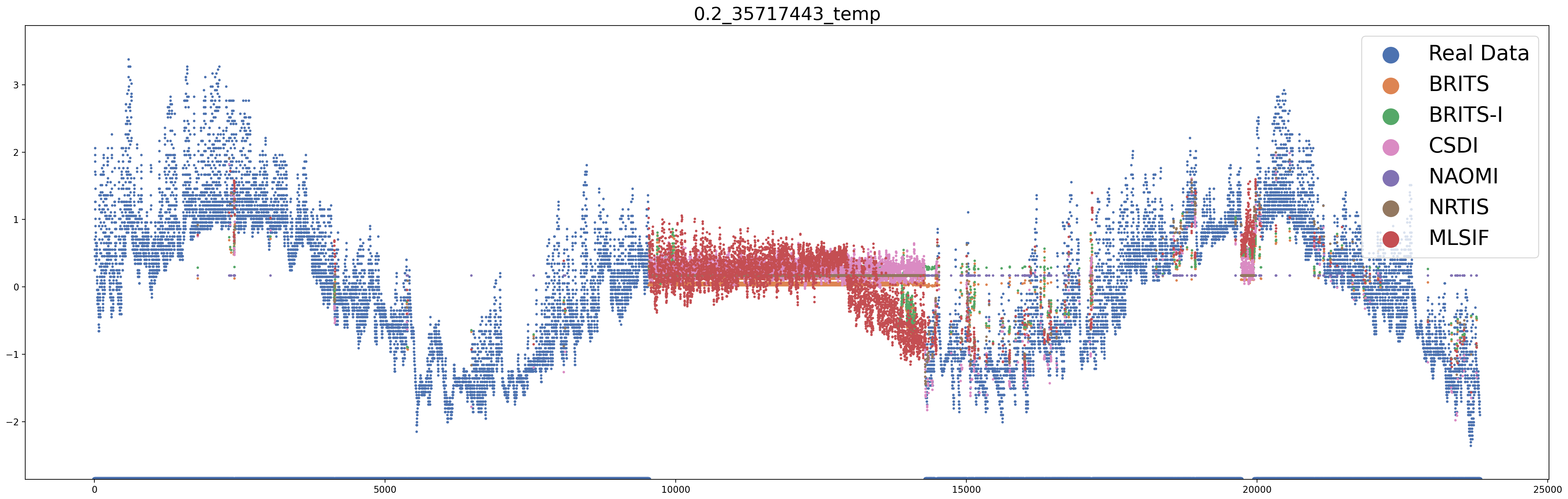}
		\end{minipage}
		\label{20 missing}
	}
	\subfloat[30\% missing]{
		\begin{minipage}[!t]{0.45\textwidth}
			\centering
			\includegraphics[width=1\textwidth]{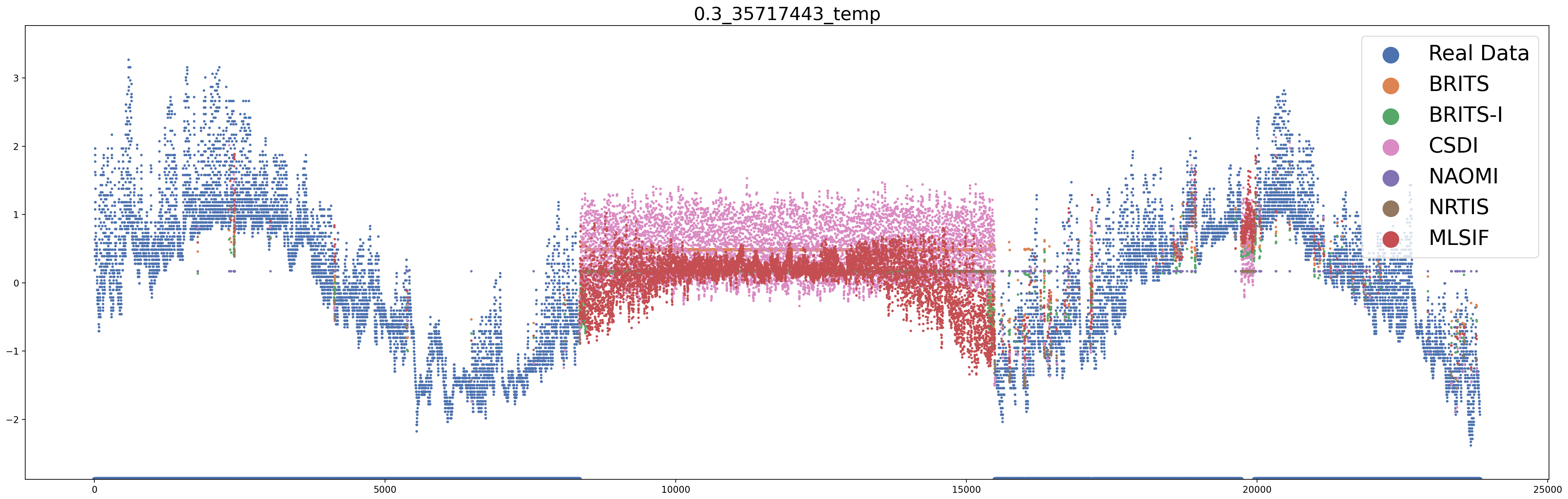}
		\end{minipage}
		\label{30 missing}
	}
	\\ 
	\subfloat[40\% missing]{
		\begin{minipage}[!t]{0.45\textwidth}
			\centering
			\includegraphics[width=1\textwidth]{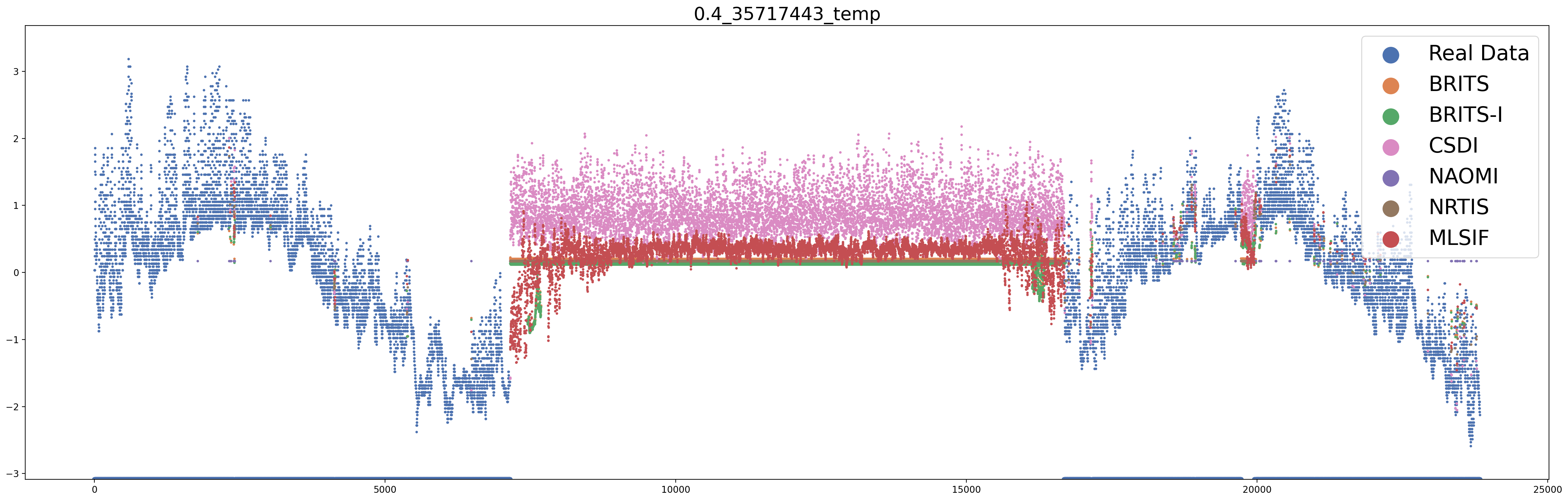}
		\end{minipage}
		\label{40 missing}
	}
	\subfloat[50\% missing]{
		\begin{minipage}[!t]{0.45\textwidth}
			\centering
			\includegraphics[width=1\textwidth]{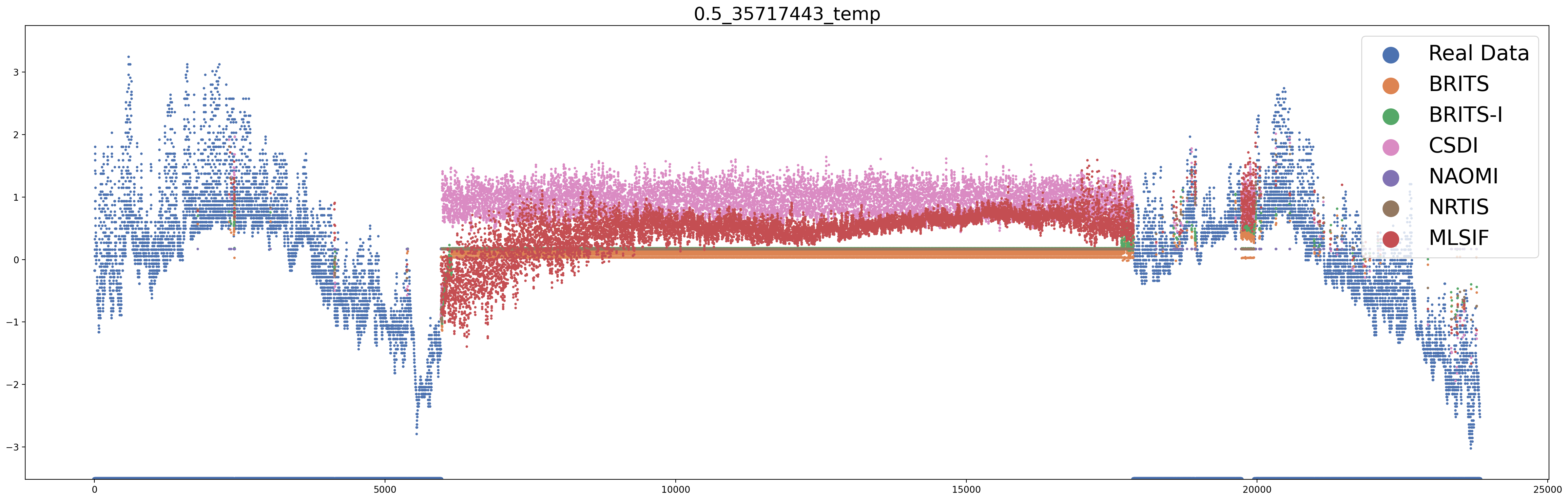}
		\end{minipage}
		\label{50 missing}
	}
	\\ 
	\caption{Imputation results of data with different missing rate using deep learning methods.}
	\label{missing rate}
\end{figure*}

The imputation results and corresponding metrics (MSE and Global SIV) are shown in Fig. \ref{missing rate} and Table \ref{missing rate table}, respectively. In Fig. \ref{missing rate}, only the results of deep learning methods in data imputation are shown due to space constraints, whereas the metrics for all methods are displayed in Table \ref{missing rate table}. 

It can be observed from Fig. \ref{missing rate} that compared with other deep learning methods, MLSIF has a good imputation ability in the face of small- and large-missing segments. Especially when dealing with large-missing segments, MLSIF can still impute the missing parts according to the known data. Other deep learning methods fail when confronted with the same problem, imputing either the same or random values. The results of the metrics in Table \ref{missing rate table} are consistent with the intuition. MLSIF can almost achieve the best results under both metrics when the missing rate is greater than 20\%. Other methods are still competitive when the missing rate is less than 10\%. It is also worth noting that, in most cases, the effect of deep learning is better than that of statistical methods in terms of indicators.

\begin{table*}[]
	\caption{MSE and Global SIV of data imputation with different missing rate using deep learning and statistics methods.}
	\centering
	\begin{tabular}{c|cc|cc|cc|cc|cc|cc}
		\hline
		\multirow{2}{*}{methods} & \multicolumn{2}{c|}{0\%} & \multicolumn{2}{c|}{10\%}         & \multicolumn{2}{c|}{20\%}         & \multicolumn{2}{c|}{30\%}         & \multicolumn{2}{c|}{40\%}         & \multicolumn{2}{c}{50\%}          \\ \cline{2-13} 
		& mse   & siv              & mse             & siv             & mse             & siv             & mse             & siv             & mse             & siv             & mse             & siv             \\ \hline
		aggregate                & -     & 0.0016           & 0.4465          & 0.7038          & 0.9901 & 0.9163          & 1.4468 & 0.8061          & 1.6181          & 0.0111 & -               & -               \\
		approx                   & -     & 0.1018           & 0.3973 & 1.1184          & 1.3879          & 0.0264 & 1.841           & 0.2198 & 1.455  & 1.4625          & 1.1858          & 1.6105 \\
		spline                   & -     & 0.0974           & 0.4134          & 1.201           & 42.9614         & 7.3364          & 13.677          & 2.041           & 1214.531        & 451.4788        & 51.2403         & 19.3022         \\
		interpolation            & -     & 0.0017  & 0.4904          & 0.0326 & 1.9098          & 0.049           & 1.8354          & 0.2663          & 1.4855          & 1.7797          & 1.1639 & 1.9565          \\
		kalman                   & -     & 0.7093           & 52.7797         & 3.9657          & 9655.416        & 1833.317        & 3.4357          & 0.2284          & 249505.5        & 100113.9        & 2615.578        & 1250.991        \\ \hline
		BRITS                    & -     & 0.0244           & \textbf{0.2806} & 1.0958          & 0.5528          & 0.0856          & 0.8542          & 1.5182          & 0.935           & 0.1286          & \textbf{1.0389} & 1.5358          \\
		BRITS-I                  & -     & 0.034            & 0.5155          & 0.6348          & 0.4463          & 1.1412          & 0.696           & 0.8921          & 0.885           & 0.1078          & 1.1024          & 0.9712          \\
		CSDI                     & -     & \textbf{0.0001}  & 0.9817          & \textbf{0.0045} & 0.4935          & 0.8739          & 0.9886          & 1.2531          & 2.1484          & 0.3234          & 2.5333          & 0.3766          \\
		NAOMI                    & -     & 0.0168           & 0.5195          & 0.3827          & 0.5224          & 0.823           & 0.7152          & 0.9769          & 0.9572          & 0.1097          & 1.1177          & 0.9732          \\
		NRTIS                    & -     & \textbf{0.0001}  & 0.5188          & 0.0707          & 0.5096          & 0.0848          & 0.7104          & 0.8733          & 0.9563          & 0.1027          & 1.1141          & 1.1545          \\
		MLSIF                    & -     & 0.0043           & 0.7134          & 0.0098          & \textbf{0.4358} & \textbf{0.0178} & \textbf{0.6015} & \textbf{0.0337} & \textbf{0.8468} & \textbf{0.0791} & 1.2143          & \textbf{0.0987} \\ \hline
	\end{tabular}
	\label{missing rate table}
\end{table*}

\subsubsection*{\bf Experiment 4}
In this experiment, we apply the model to three real geological sensor datasets with missing values (45710421\_x, 45710421\_y, and 45710422\_x). There are approximately 20,000 timestamps in each dataset used in this experiment, with a missing rate greater than 30\%. 

\begin{figure*}[!t]
	\centering
	\subfloat[45710421\_x imputed by R]{
		\begin{minipage}[!t]{0.45\textwidth}
			\centering
			\includegraphics[width=1\textwidth]{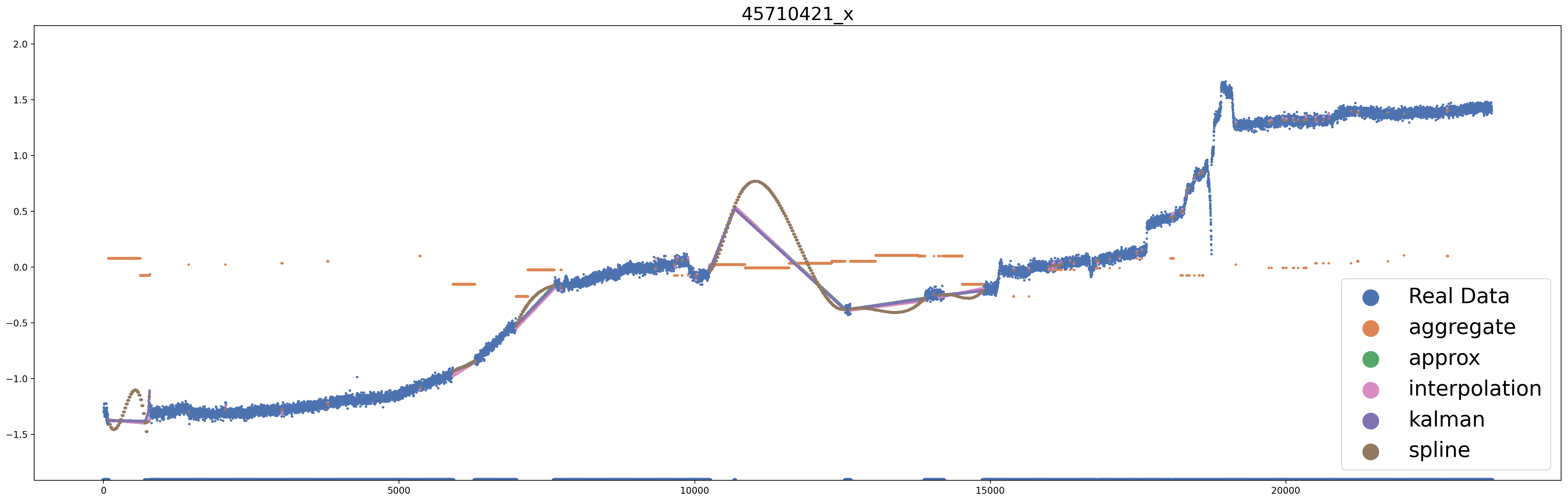}
		\end{minipage}
	}
	\subfloat[45710421\_x imputed by deep learning methods]{
		\begin{minipage}[!t]{0.45\textwidth}
			\centering
			\includegraphics[width=1\textwidth]{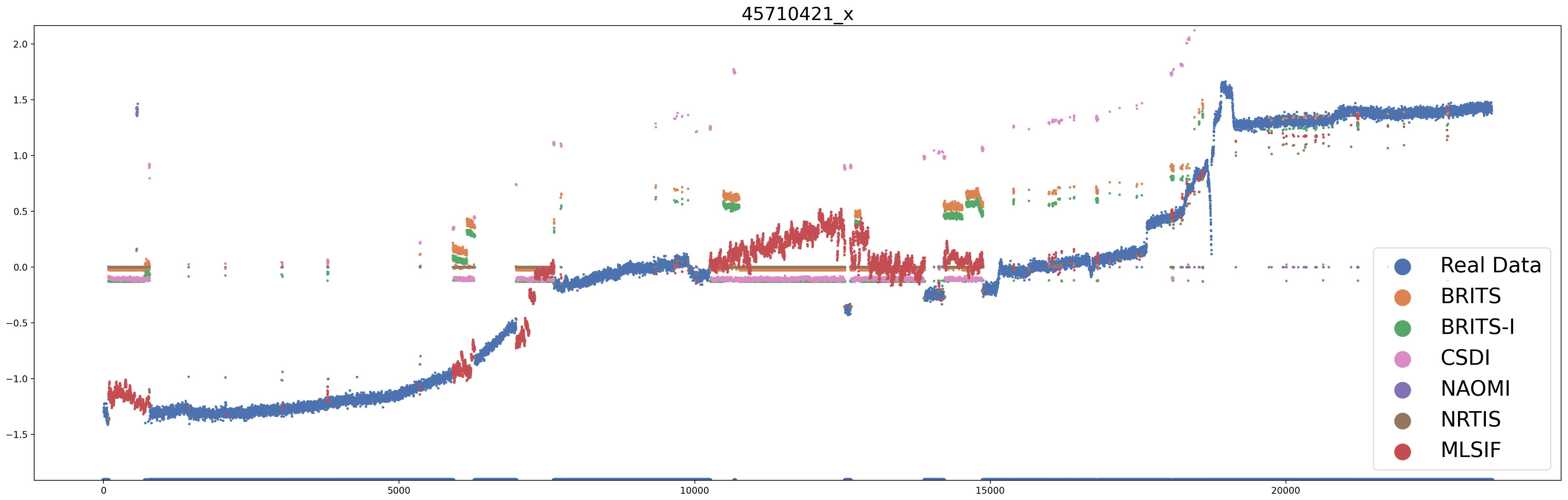}
		\end{minipage}
	}
	\\ 
	\subfloat[45710421\_y imputed by R]{
		\begin{minipage}[!t]{0.45\textwidth}
			\centering
			\includegraphics[width=1\textwidth]{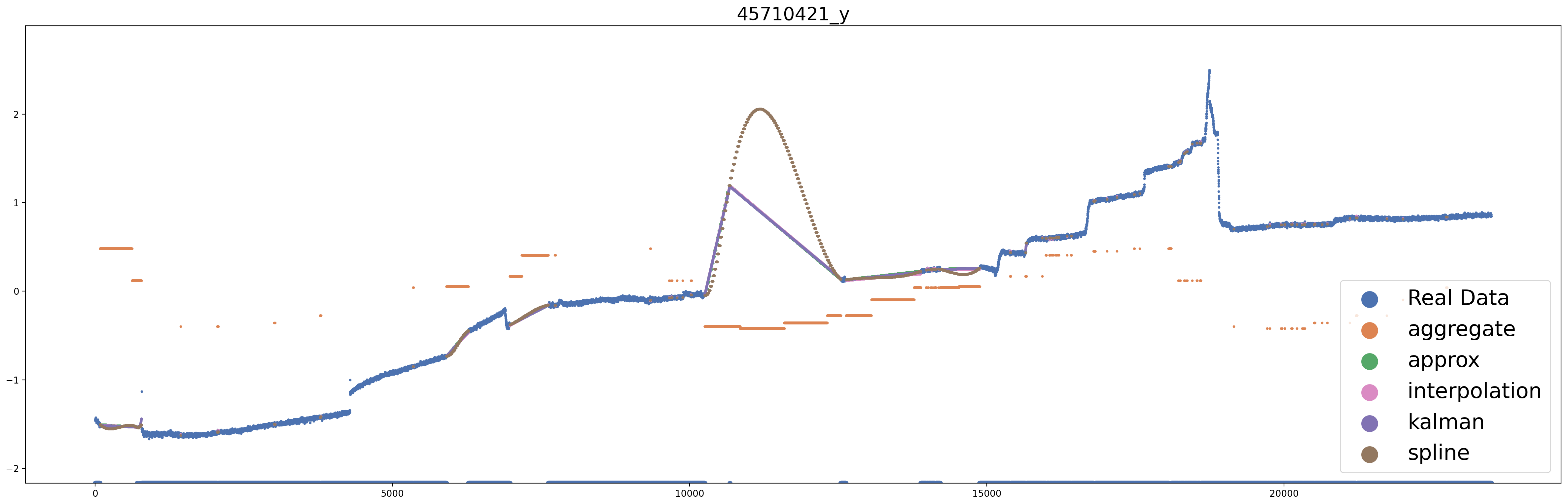}
		\end{minipage}
	}
	\subfloat[45710421\_y imputed by deep learning methods]{
		\begin{minipage}[!t]{0.45\textwidth}
			\centering
			\includegraphics[width=1\textwidth]{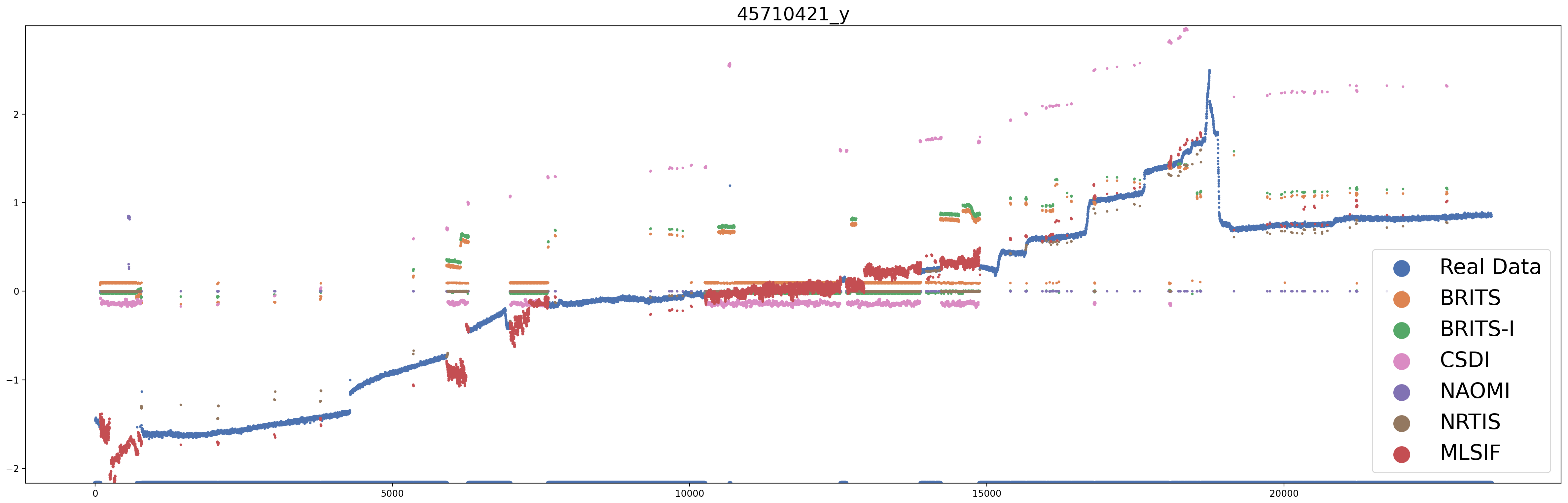}
		\end{minipage}
	}
	\\ 
	\subfloat[45710422\_x imputed by R]{
		\begin{minipage}[!t]{0.45\textwidth}
			\centering
			\includegraphics[width=1\textwidth]{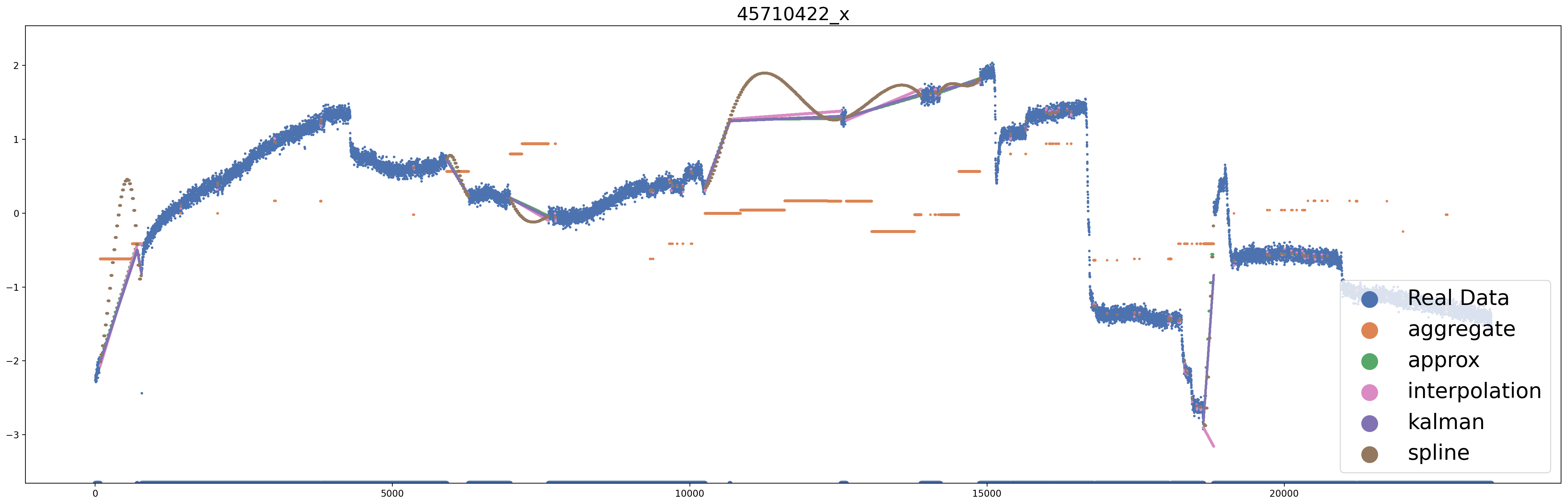}
		\end{minipage}
	}
	\subfloat[45710422\_x imputed by deep learning methods]{
		\begin{minipage}[!t]{0.45\textwidth}
			\centering
			\includegraphics[width=1\textwidth]{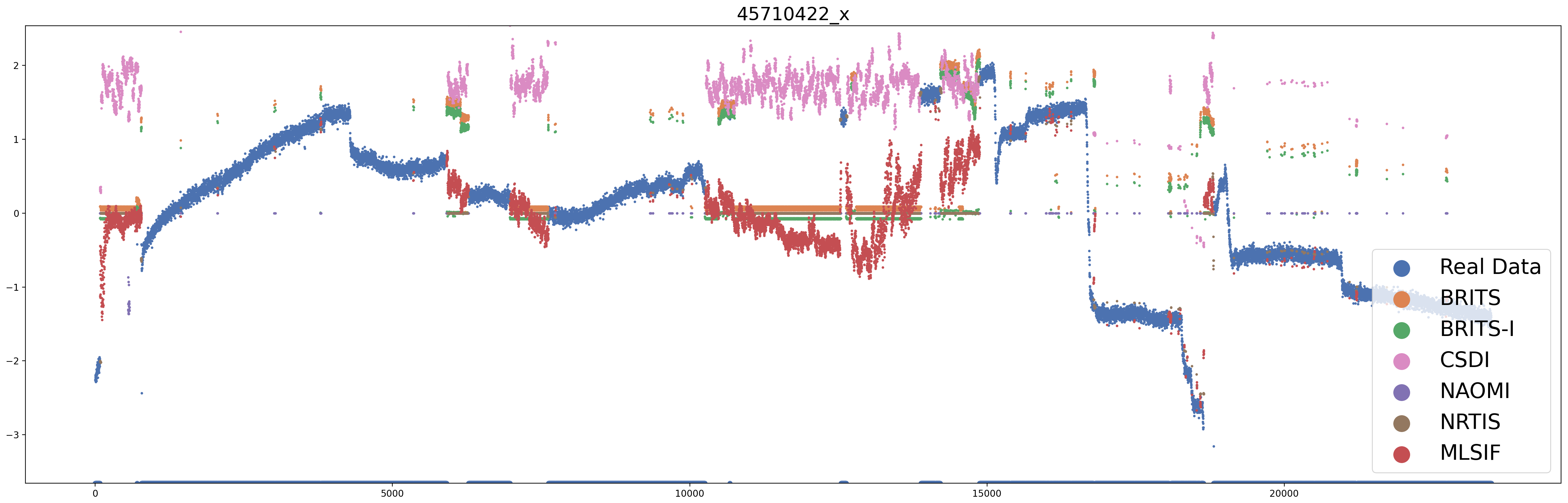}
		\end{minipage}
	}
	\\ 
	\caption{The imputation results of Experiment 3 by R package and deep learning methods. The blue on the abscissa represents the coordinate corresponding to the location where the observed value exists.}
	\label{experment 3}
\end{figure*}

The experimental results are shown in Fig. \ref{experment 3}. From the overall point of view of the data, the imputation results of the statistical methods are very good when small segments are missing, almost indistinguishable from the original data. However, traces of imputation can be seen in the missing positions of large segments. When there are no real values in a sample, the model training becomes out of control, which is a problem for these deep learning methods. This type of sample is removed during training. The results show that while the validation loss is decreasing (low to 0.01 level), the imputation results are not improving. As in the previous experiment, when faced with a large segment of missing values, the results of imputation using other deep learning methods are either the same value or a random value around a certain value. 

The MLSIF can not only impute good results for small missing gaps but also successfully impute large-missing gaps. The MLSIF imputation results are almost consistent with the original data distribution in the positions where small segments are missing. The imputation results are nearly identical to the original data trend and adhere to the data distribution in the large missing gap position in the middle. 

\begin{table}[]
	\caption{Global SIV metric results of three geosensor data. }
	\centering
	\begin{tabular}{c|ccc}
		\hline
		\multicolumn{1}{c|}{methods} & \multicolumn{1}{c}{45710421\_x} & \multicolumn{1}{c}{45710421\_y} & \multicolumn{1}{c}{45710422\_x} \\ \hline
		aggregate                    & 0.02761                         & 0.02201                         & 0.02148                \\
		approx                       & 0.03271                         & 0.01106                         & 0.05718                         \\
		spline                       & 0.02346                & \textbf{0.01013}                & 0.07715                         \\
		interpolation                & 0.03365                         & 0.01066                         & 0.07081                         \\
		kalman                       & 0.03412                         & 0.01099                         & 0.05648                         \\ \hline
		BRITS                        & 1.00396                         & 3.07141                         & 4.08062                         \\
		BRITS-I                      & 0.93489                         & 2.97145                         & 3.77598                         \\
		CSDI                         & 0.98591                         & 2.72885                         & 5.75567                         \\
		NAOMI                        & 0.02827                         & 0.03034                         & 0.03183                         \\
		NRTIS                        & 0.02691                         & 0.02758                         & 0.02592                         \\
		MLSIF                        & \textbf{0.01474}                & 0.01048                & \textbf{0.01691}                \\ \hline
	\end{tabular}
	\label{experment 3 table}
\end{table}

Table \ref{experment 3 table} shows the Global SIV based on three geosensor data. As these datasets contain many missing values, removing regret affects the data's integrity, which also causes MSE, MAE, $R^2$, and $d_2$ from being calculated. Local SIV is excluded from the table because only MLSIF can calculate this metric. 


The application of MLSIF to practical problems is reflected in Experiment 4. It has been discovered that many models fail to produce good imputation results when faced with significant missing data gaps; however, MLSIF can handle this problem and produce good results. 

\subsection{Computational Efficiency Analysis}
Contrary to the one train-and-impute framework, MLSIF requires many cycles of training and imputing. Taking the six datasets in Experiment 3 as an example, they must go through 12, 59, 108, 151, 200 and 246 cycles of training and imputing to obtain the results. Compared to the single-stage model, its time will increase as the cycles increase. In practical, the following tricks will help save time:
\begin{itemize}
	\item{For models that are not sensitive to sequence length, such as NRTIS \cite{shan2021nrtsi}, a certain number of training iterations can be reduced by inheriting the parameters of the previous stage in each stage; (The measure taken in this paper)}
	\item{For models that are sensitive to sequence length, such as NAOMI \cite{liu2019naomi}, one can reduce a certain number of training iterations by inheriting the same input length model parameters.}
\end{itemize}

Generally, statistical methods are much more computationally efficient than deep learning methods; however, they sacrifice a certain level of accuracy. In contrast, deep learning methods trade time for precision, whereas MLSIF speeds up more time for higher precision.

\section{Conclusion}
This study proposes SIV and MLSIF for sensor data missing value imputation. The introduction of SIV loss improves the imputation models, and the SIV metric measures the imputation effect effectiveness. MLSIF uses the multistage imputation method, which uses the imputation results of previous stages as knowledge to facilitate the learning of the model and improve the effect of imputation. During the imputation process, the framework dynamically adjusts the data length according to the unimputed data. This approach can be used to adaptively deal with different degrees of missing tasks. In the experimental design, we jumped out of the inherent assumptions, by simulating the actual situation of the real missing data, to avoid the phenomenon that the verification process deviates from reality. This paper only discusses one-dimensional sensor data. Extending the proposed method to multidimensional data imputation should be researched in future studies. Additionally, more measurement methods based on missing values must be explored. 


\bibliographystyle{IEEEtran}
\bibliography{MLSIF}

\newpage

\section{Biography Section}

\begin{IEEEbiography}[{\includegraphics[width=1in,height=1.25in,clip,keepaspectratio]{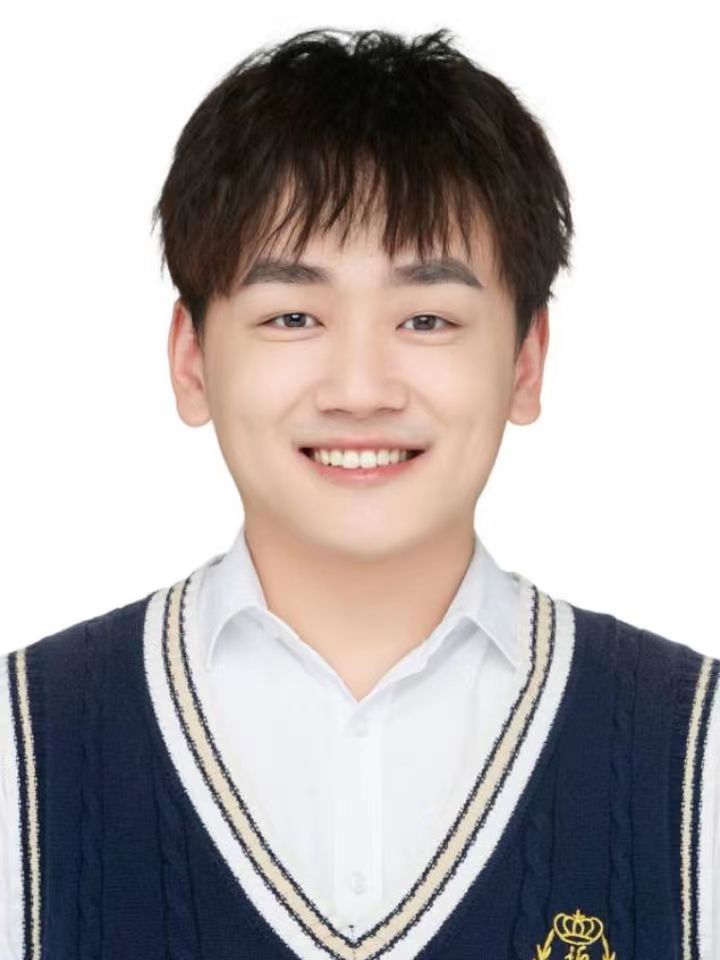}}]{Jin-Sheng Yang}
	received a bachelor's degree from Sichuan University in 2016. Currently, he is a postgraduate student at Hainan University. His main research direction is time series data analysis.
\end{IEEEbiography}

\begin{IEEEbiography}[{\includegraphics[width=1in,height=1.25in,clip,keepaspectratio]{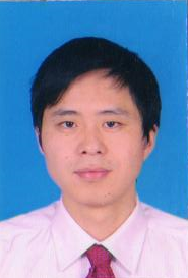}}]{Yuan-Hai Shao}
	received his B.S. degree in information and computing science from College of Mathematics, Jilin University, a master's degree in applied mathematics, and a Ph.D. degree in operations research and management in College of Science from China Agricultural University, China, in 2006, 2008 and 2011, respectively. Currently, he is a Full Professor at the School of Management, Hainan University, Haikou, China. His research interests include support vector machines, optimization methods, machine learning and data mining. He has published over 100 refereed papers on these areas, including IEEE TPAMI, IEEE TNNLS, IEEE TC, PR, and NN.
\end{IEEEbiography}

\begin{IEEEbiography}[{\includegraphics[width=1in,height=1.25in,clip,keepaspectratio]{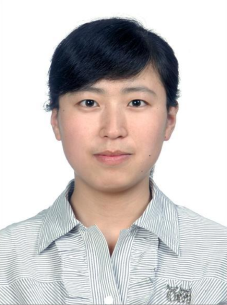}}]{Chun-Na Li}
	received her Master's degree and Ph.D degree in Department of Mathematics from Harbin Institute of Technology, China, in 2009 and 2012, respectively. Currently, she is a professor at Management School, Hainan University. Her research interests include optimization methods, machine learning and data mining.
\end{IEEEbiography}

\begin{IEEEbiography}[{\includegraphics[width=1in,height=1.25in,clip,keepaspectratio]{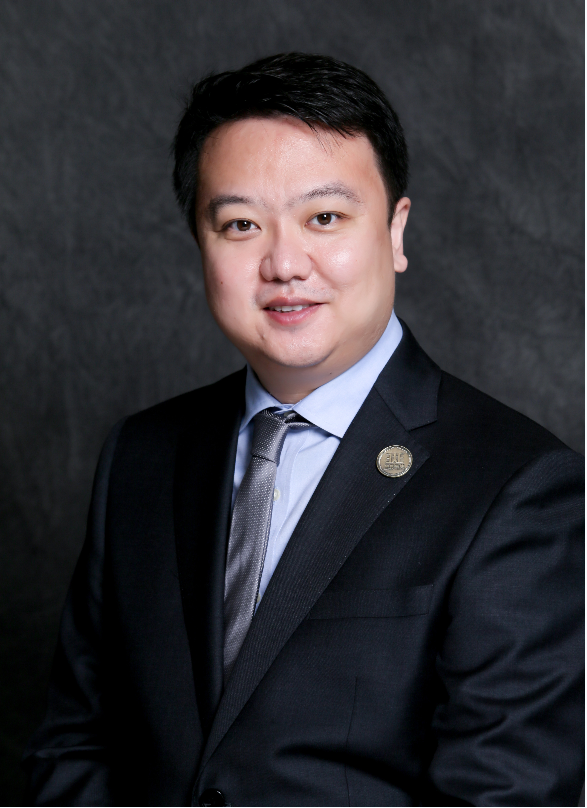}}]{Wen-si Wang}
	received his Master’s and Ph.D. degrees in Microelectronics from Tyndall National Institute, Republic of Ireland, in 2008 and 2012, respectively. He was a visiting scholar with the Georgia Institute of Technology, Atlanta in 2012. From 2013 to 2015, he was with Tyndall National Institute as Post-doc and Assistant Researcher. Since 2015, he has been with the Beijing University of Technology as an Associate Professor. He is also the co-founder of a medical R\&D company SuperVision with its focus on A.I. in medical applications. He has published over 30 papers and filed over 20 patents in this area.   
\end{IEEEbiography}

\vfill

\end{document}